\documentclass{article}

\usepackage{microtype}
\usepackage{graphicx}
\usepackage{booktabs} 
\usepackage{svg} 
\usepackage{subfig}
\usepackage[ruled,vlined]{algorithm2e}

\usepackage{hyperref}

\usepackage[accepted]{icml2023}

\usepackage{amsmath}
\usepackage{amssymb}
\usepackage{mathtools}
\usepackage{amsthm}

\usepackage[capitalize,noabbrev]{cleveref}

\theoremstyle{plain}

\theoremstyle{definition}

\theoremstyle{remark}

\usepackage[textsize=tiny]{todonotes}

\icmltitlerunning{dugMatting: Decomposed-Uncertainty-Guided Matting}

\begin{document}

\twocolumn[
           
\icmltitle{dugMatting: Decomposed-Uncertainty-Guided Matting}




\begin{icmlauthorlist}
\icmlauthor{Jiawei Wu}{w}
\icmlauthor{Changqing Zhang}{z}
\icmlauthor{Zuoyong Li}{l}
\icmlauthor{Huazhu Fu}{f}
\icmlauthor{Xi Peng}{p}
\icmlauthor{Joey Tianyi Zhou}{f,zz}
\end{icmlauthorlist}

\icmlaffiliation{w}{College of Mechanical and Electrical Engineering, Fujian Agriculture and Forestry University, Fuzhou, China}
\icmlaffiliation{z}{College of Intelligence and Computing, Tianjin University, Tianjin, China}
\icmlaffiliation{l}{Fujian Provincial Key Laboratory of Information Processing and Intelligent Control, Minjiang University, Fuzhou, China}
\icmlaffiliation{f}{Institute of High Performance Computing, Agency for Science, Technology and Research, Singapore}
\icmlaffiliation{p}{College of Computer Science, Sichuan University, Chengdu, China}
\icmlaffiliation{zz}{Centre for Frontier AI Research (CFAR), Agency for Science, Technology and Research (A*STAR), Singapore}

\icmlcorrespondingauthor{Changqing Zhang}{zhangchangqing@tju.edu.cn}
\icmlcorrespondingauthor{Zuoyong Li}{fzulzytdq@126.com}

\icmlkeywords{Machine Learning, ICML}

\vskip 0.3in
]



\printAffiliationsAndNotice{} 

\begin{abstract}
Cutting out an object and estimating its opacity mask, known as image matting, is a key task in image and video editing. Due to the highly ill-posed issue, additional inputs, typically user-defined trimaps or scribbles, are usually needed to reduce the uncertainty. Although effective, it is either time consuming or only suitable for experienced users who know where to place the strokes. In this work, we propose a decomposed-uncertainty-guided matting (dugMatting) algorithm, which explores the explicitly decomposed uncertainties to efficiently and effectively improve the results. Basing on the characteristic of these uncertainties, the epistemic uncertainty is reduced in the process of guiding interaction (which introduces prior knowledge), while the aleatoric uncertainty is reduced in modeling data distribution (which introduces statistics for both data and possible noise). The proposed matting framework relieves the requirement for users to determine the interaction areas by using simple and efficient labeling. Extensively quantitative and qualitative results validate that the proposed method significantly improves the original matting algorithms in terms of both efficiency and efficacy.
\end{abstract}


\section{Introduction}
\label{sec:intro}

Digital image matting is the estimation of the opacity of foreground or background from an image, which is one of the fundamental elements in many applications, e.g., compositing live-action and rendered elements together, and performing local color corrections. Specifically, given an image $I$, image matting can be regarded as a linear combination of foreground $F \in \mathbb{R}^{H \times W \times C}$ and background $B \in \mathbb{R}^{H \times W \times C}$ with the alpha matte $\mu \in [0,1]^{H\times W}$ as follows:
\begin{equation}
    I_m = \mu_m F_m + (1-\mu_m) B_m, \nonumber
\end{equation}
where $m=(x,y)$ denotes the pixel position.

Since the estimation of $\mu$ without any extra information is a highly ill-posed problem, traditional algorithms \cite{re:ACFS, re:KM, re:alphaM, re:DIM, re:GCAM, re:LFP, re:matteFormer} usually introduce a trimap to confine the solution space. The trimap separates a picture into two known foreground and background regions along with an unknown transition region. Hence, the matting task is simplified as the problem of estimating the opacity in the transition region. Based on this simplification, the recently proposed matteformer \cite{re:matteFormer} achieves the state-of-the-art performance. However, drawing a suitable trimap is still time-consuming and tedious. For some complex cases, it will even cost more than 10 minutes \cite{re:improveM}.

Recently, some trimap-free matting algorithms attempt to eliminate the model dependence on the prior labeling. However, the performance of trimap-free methods \cite{re:gfm, re:modnet, re:u2net, re:SHM, re:P3M} still lags far behind the trimap-based methods. The inherent reason is that these models cannot determine which foreground target should be extracted without the guidance of trimap. Therefore, existing trimap-free methods are only able to extract the class-specific objects (e.g., portrait, animal) or salient objects after training on large-scale matting data. Moreover, trimap-free methods is powerless when users want to choose a new category. To balance the efficiency and effectiveness, some novel interactive strategies have been introduced for matting. With user scribbles or clicks, interactive matting achieves similar performance to the trimap-based approaches in relatively low labeling cost~\cite{re:improveM}. However, a promising outcome usually requires multiple interactions because the interactions heavily rely on user experience, leading to long-term interaction (the shortest click interaction method still takes about 20 seconds~\cite{re:improveM}). Besides, the matting performance may be unstable due to the ambiguity of the user interaction.

\begin{figure}[t]
    \centering
    \includegraphics[width=\linewidth]{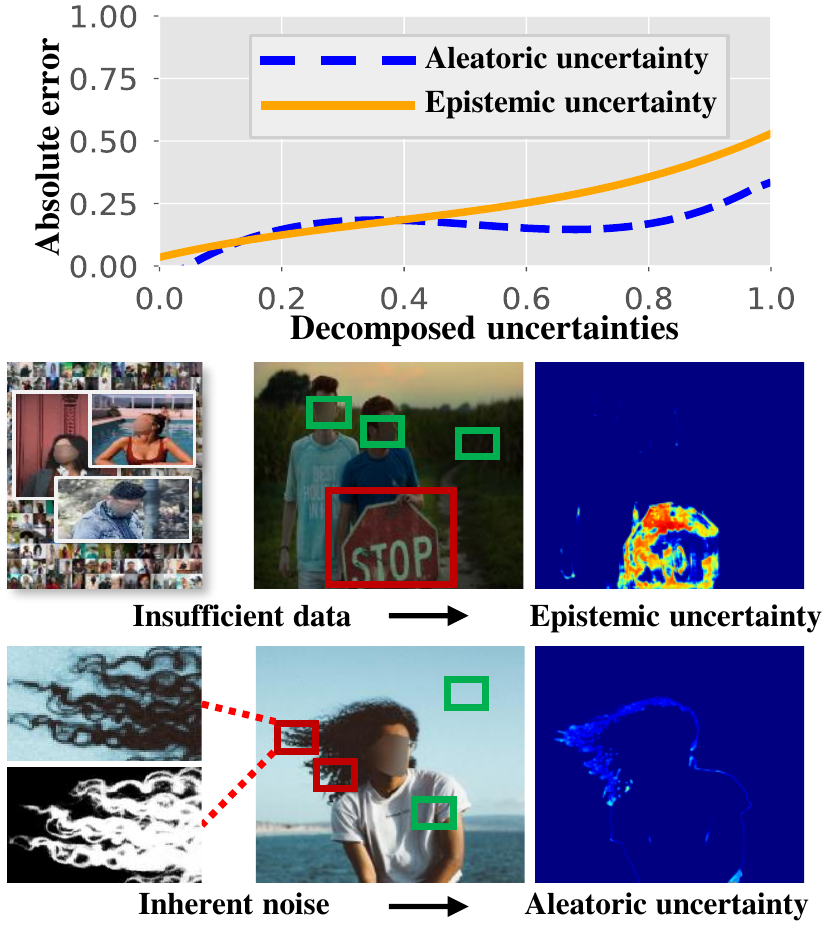}
    \caption{Motivation of the proposed dugMatting. The matting performance could be significantly improved by reducing the decomposed epistemic and aleatoric uncertainties (top row), where these uncertainties are ubiquitous in learning-based image matting (middle and bottom rows). The epistemic uncertainty cannot be reduced by model but can be reduced by user interaction, while the aleatoric uncertainty is difficult to be reduced by human but can be reduced by handling the data noise. Therefore, it is attractive to decompose these uncertainties and exploit them accordingly.}
    \label{fig:intro}
\end{figure}

To relieve the restriction of user experience, we propose a decomposed-uncertainty-guided matting (dugMatting) algorithm, which elegantly exploits the decomposed epistemic and aleatoric uncertainties. As shown in~\cref{fig:intro}, the epistemic uncertainty basically results from insufficient training data while the aleatoric uncertainty often appears in transition regions due to the inherent noise. Based on the observation that the absolute error is highly correlated with the these uncertainties, a natural question is \textit{can we effectively reduce the decomposed uncertainty}?

{\textbf{Contribution.} Epistemic uncertainty is often due to a lack of training data and thus it is difficult to be reduced by models themselves, while it can be reduced by interaction. Aleatoric uncertainty refers to the uncertainty inherent in the observations, e.g., measurement noise or inaccurate labeling, which is more intricate but can be reduced by handling possible noise, e.g., using data augmentation~\cite{re:aleaAug, re:reduceAl}. We propose a decomposed-uncertainty-guided matting framework, where the epistemic uncertainty \cite{re:WN, re:DVR} is used to identify proposal regions for user interaction. Accordingly, users only need labeling these regions. To reduce the aleatoric uncertainty, a plug-and-play module based on the estimated data distribution is devised where the augmentation is realized. Specifically, we model the matting output as a Normal-Inverse-Gamma distribution, which hierarchically characterizes the uncertainties and accordingly promotes both regression accuracy and trustworthiness \cite{re:DVR}. Different from the standard setting, the Normal-Inverse-Gamma distribution depends on both the input image and interaction. Therefore, multiple interactions on an image yield multiple NIG distributions, where we introduce NIG summation \cite{re:TMR, re:BBLR} to combine these multiple NIG distributions improving the stability. The contributions of this work are summarized as follows:
\vspace{-10px}

\begin{itemize}
    \item For the first time, we reveal the relationship between epistemic/aleatoric uncertainties and the matting error, and thus transform the matting promotion into the problem of epistemic/aleatoric uncertainties reduction.\vspace{-5px}
    \item We propose a decomposed-uncertainty-guided matting algorithm, where the epistemic uncertainty is utilized to actively provide interaction proposals for users and the aleatoric uncertainty is used to guide the matte refinement in a plug-and-play module.\vspace{-5px}
    \item We conduct extensive experiments on multiple real-world benchmarks, which demonstrate that the proposed method not only improves the performance of trimap-based matting, but also enables trimap-free matting to extract novel foreground. 
\end{itemize}

\section{Related Work}
\subsection{Image Matting}
Image matting refers to extracting interesting foreground or background with fine details from an image, which can be divided into prior-based matting \cite{re:ACFS,re:alphaM,re:DIM,re:MGMatting,re:matteFormer} and prior-free matting \cite{re:gfm, re:modnet, re:SHM, re:P3M, re:u2net}. The prior-based matting methods require an additional prior for constraining the solution space. One typical trimap separates an image into foreground, background, and transition regions, where only the opacity of transition regions is unknown. Before the deep learning period, some well-established methods \cite{re:LBDM,re:KM,re:ACFS,re:RWM,re:BayeM,re:cluM,re:GSM} solve the matting task based on trimap prior. For example, the closed-form matting \cite{re:ACFS} derives a cost function based on local smoothing of foreground and background colors, and the globally optimal alpha matting is accordingly induced by solving a sparse system of linear equations. In the era of deep learning, data-driven methods have emerged in matting community, exhibiting much better performance than conventional methods. For example, deep image matting (DIM)~\cite{re:DIM} uses a convolutional network to refine the alpha matte predicted under the encoder-decoder framework, allowing for higher accuracy and sharper edges. A guided contextual attention block is designed in GCANet~\cite{re:GCAM} to integrate the alpha stream information and image information, and improve the details of matting as well. LPFNet~\cite{re:LFP} models the long-range context features outside the reception fields to improve the alpha matte results. To relieve the load in manually constructing a trimap, the prior-free methods often divide the matting task into a triamp generation and a trimap-based matting subtasks~\cite{re:gfm}. However, these trimap-free methods fail to handle arbitrary foreground due to the model ambiguity without guidance. 
\begin{figure*}[t]
    \centering
    \includegraphics[width=\linewidth]{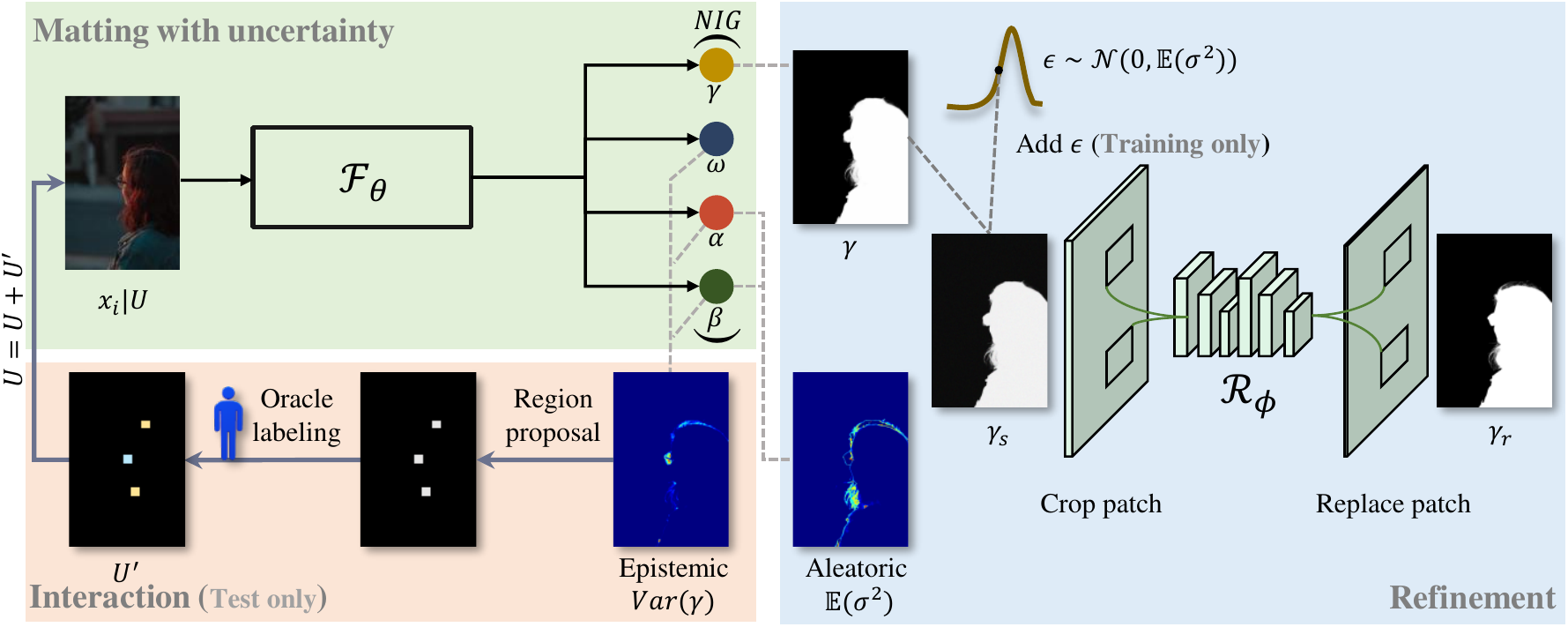}
    \caption{Illustration of the proposed decomposed-uncertainty-guided matting framework. The matting network fits a NIG distribution, proposing interactive regions for user based on epistemic uncertainty and detail regions for refined module based on aleatoric uncertainty.}
    \label{fig:framework}
\end{figure*}

\subsection{Uncertainty Estimation}
Uncertainty estimation in deep networks has attracted significant attention \cite{re:ensembleUn, re:MCDropout, re:WN, re:DVR, re:EDC, re:ICMLun1, re:ICML_un2}, especially when the systems are deployed in safety-critical tasks such as autonomous car control and medical diagnosis. Basically, uncertainty can be roughly divided into aleatoric uncertainty and epistemic uncertainty, in which aleatoric uncertainty captures noise inherent in the observations and epistemic uncertainty captures our ignorance about which model generated our collected data \cite{re:WN}. For modeling aleatoric uncertainty, the network often outputs a Gaussian distribution with a learnable variance. For modeling epistemic uncertainty, Bayesian-based methods \cite{re:bayeTUN,re:bayeUN1,re:BayeIUN, re:ICML_un3} form a predictive distribution by marginalizing the distribution over model parameters. To reduce the computation of Bayesian network, dropout or ensemble are used to approximate variational Bayesian inference~\cite{re:ensembleUn, re:MCDropout}, but these methods require multiple forwards. In contrast, some models directly predict the parameters of conjugate prior distribution on the predicted target distribution. Then, one forward pass can estimate the target and the associated uncertainty. Most of those models focus on classification and thus usually estimate the parameters of a Dirichlet distribution~\cite{re:DIR1,re:EDC,re:DIR2,re:DIR3,re:DIR4}. Since image matting is an intrinsically regression problem, we introduce a Normal-Inverse-Gamma (NIG) distribution~\cite{re:eva_epis} to characterize the uncertainty.

\section{Proposed Method}
\subsection{Preliminary of Evidence-based Uncertainty}
\label{sec:NIG}

We briefly introduce the regression under the evidence-based uncertainty estimation. Regression task can be solved from a maximum likelihood perspective with Gaussian distribution. Given the training data $\mathcal D=\{x_i, y_i\}^N_{i=1}$, maximum likelihood estimation (MLE) is achieved by minimizing the negative log likelihood loss function
\begin{equation}
    \mathcal L_i(\theta) = \frac{(y_i-\mu)^2}{2\sigma^2} + \log\sigma, \nonumber
    \label{eq:MLE}
\end{equation}
where $\theta$ denotes the parameters of matting network, $\mu$ and $\sigma$ denote the mean and variance parameters of Gaussian distribution respectively, which are typically learned through deep neural networks. Existing matting networks target at learning the alpha matte (mean $\mu$) only. When $\mu$ and $\sigma$ are all learnable, the likelihood function successfully models the aleatoric uncertainty (variance), also known as the data uncertainty. However, epistemic uncertainty, also known as model uncertainty, often requires additional estimation based on the Bayesian framework, e.g., MC Dropout~\cite{re:MCDropout} and ensemble~\cite{re:ensembleUn}.

To jointly model aleatoric and epistemic uncertainties, the mean $\mu$ and variance $\sigma^2$ are assumed to be drawn from Gaussian and Inverse-Gamma distributions, respectively. Then the Normal Inverse-Gamma (NIG) distribution $  \text{NIG}(\gamma,\omega, \alpha, \beta)$ can be considered as a higher-order conjugate prior of the Gaussian distribution
\begin{equation}
\begin{split}
    y_i \sim &\mathcal N(\mu, \sigma),\\ 
    \mu \sim \mathcal N(\gamma, \sigma^2\omega^{-1})&, \qquad \sigma^2\sim \Gamma^{-1}(\alpha, \beta), \nonumber
\end{split}
\end{equation}
where $\Gamma(\cdot)$ denotes the gamma function. In this case, the distribution of $y$ takes the form of a NIG$(\gamma,\omega, \alpha, \beta)$ distribution
\begin{equation}
\begin{split}
    p(\mu, \sigma | \gamma,\omega, \alpha, \beta) = \frac{\beta^\alpha}{\Gamma(\alpha)}&\frac{\sqrt{\omega}}{\sigma\sqrt{2\pi}}(\frac{1}{\sigma^2})^{\alpha+1} \\ & \exp{-\frac{2\beta+\omega(\sigma-\mu)^2}{2\sigma^2}}, \nonumber
\end{split}
\end{equation}
where $\gamma \in R$, $\omega>0$, $\alpha>1$ and $\beta > 0$. The total evidence is the sum of all vitual-observations counts $2 \omega+\alpha$. To solve the NIG distribution during training phase, the following loss \cite{re:DVR} is induced to minimize the negative log likelihood
\begin{equation}
\begin{split}
    \mathcal L^{NLL}(\theta) = \frac{1}{2}&\log(\frac{\pi}{\omega}) - \alpha\log(\Omega) + \\ (\alpha&+\frac{1}{2})\log((y-\gamma)^2\omega+\Omega)+\log\Phi,
    \label{eq:NLL}
\end{split}
\end{equation}
where $\Omega=2\beta(1+\omega)$ and $\Phi=\left(\frac{\Gamma(\alpha)}{\Gamma(\alpha+\frac{1}{2})}\right)$. To further constrain the incorrect evidence, a regularizer is introduced in the total loss
\begin{equation}
    \mathcal L_{NIG}(\theta)=\mathcal L^{NLL}(\theta) + \lambda\mathcal{L}^R(\theta), \nonumber
    \label{eq:NIG_loss}
\end{equation}
where $\mathcal L^R(\theta)=|y_i - \gamma| \cdot (2\omega + \alpha)$ is the penalty for incorrect evidence, and the coefficient $\lambda>0$ balances these two loss terms.

\subsection{Integrating Uncertainty into Matting}
Image matting can be considered as a regression task, where the output is the alpha matte $\mu \in [0,1]$ conditioning on the user map $U$
\begin{equation}
    \mu = \mathcal F_\theta(x_i|U), \nonumber
\end{equation}
where $\mathcal F_\theta$ denotes the matting network, and the user map  $U$ is empty for trimap-free matting. In order to characterize uncertainty for existing matting networks, we propose to replace the deterministic output with a NIG distribution following \cref{sec:NIG}
\begin{equation}
    NIG(\gamma,\omega, \alpha, \beta) = \mathcal F_\theta(x_i|U), \nonumber
\end{equation}
where $\gamma \in [0,1]$, $\omega>0$, $\alpha>1$, and $\beta>0$. Specifically, we first extend the last layer of matting network to output $\gamma,\omega, \alpha, \beta$ by four independent linear layers with shared features as shown in \cref{fig:framework}. Then, we apply activation functions $sigmoid, softplus, softplus+1, softplus$ for $\gamma,\omega, \alpha, \beta$ to ensure the proper ranges. Although simple, the modification well fits most existing matting networks. Accordingly, the aleatoric and epistemic uncertainties are obtained as
\begin{equation}
   \underbrace{ \mathbb{E}[\sigma^2]=\frac{\beta}{\alpha-1}}_{aleatoric}, \quad \underbrace{Var[\gamma]=\frac{\beta}{\omega(\alpha-1)}}_{epistemic}. \nonumber
   \label{eq:uncertainty}
\end{equation}


\begin{algorithm}[htb]
\caption{Uncertainty-Guided Interaction.}
\label{alg:Framwork} 
\SetAlgoLined
\KwIn{Epistemic uncertainty $u_{epis}$, predicted matte $\gamma$, input image $x$, threshold $t$, patch number $K$, and selection number $N$.}
\textbf{Initialization:} Initialize the user map $U$. \\ 
    Divide ${u_{epis}}$ into $K \times K$ patches. \\
    Compute the patch-level uncertainty $u_p \in \mathbb R_+^{K \times K}$.\\
    $\mathcal P \leftarrow$ Top $N$ uncertainty patches from $\{u_p | u_p > t\}$. \\
    \For {$p$ in $\mathcal P$}{
        \hspace{0.2px}Calculate the index $I$ of $p$ in input image $x$. \\
        Users select a label from foreground, background, or transition for $x[I]$. \\
        Update user map $U$.
    }
\KwOut{User map $U$.}
\end{algorithm}

\begin{figure}[t]
    \centering
    \includegraphics[width=\linewidth]{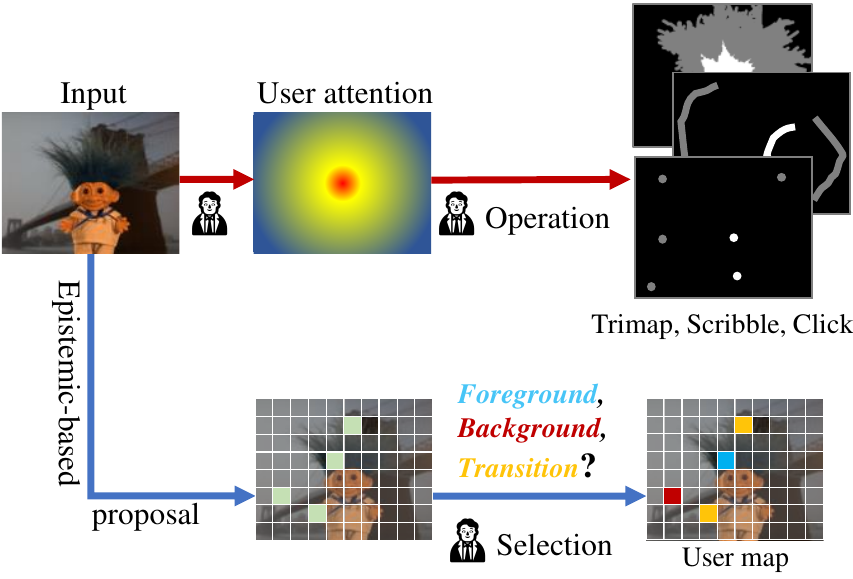}
    \caption{The proposed interaction allows the users to focus on selection.}
    \label{fig:interaction}
\end{figure}

\subsection{Epistemic Uncertainty-based Interaction}
\label{section:e_in}

Traditional interaction \cite{re:improveM,re:inerM2} implicitly contains two steps: users first empirically locate the interacted regions and then conduct interactive operation by trimap, scribble, or click. In our method, we estimate the epistemic uncertainty to automatically determine the interaction regions and then users could only select labels (foreground, background, or transition) for them as shown in Figure \ref{fig:interaction}. This novel interaction avoids the time-consuming region searching. Specifically, we divide the epistemic uncertainty map into $K \times K$ patches, where the patch-level epistemic uncertainty is the average on all pixels in each patch. Then, the proposal patch set for interaction is constructed satisfying two conditions: top $N$ patch-level epistemic uncertainty and greater than a threshold $t$. Finally, users select a label for each proposal patch. The simplified interactive process is summarized in \cref{alg:Framwork}.


To incorporate the results in previous interactions for stabilization, a direct way is to integrate the corresponding NIG distributions into a uniform one. A natural way is using the following additive way
\begin{equation}
    NIG(\gamma, \omega, \alpha, \beta)=\frac{1}{M}\sum_{m=1}^M NIG(\gamma_m, \omega_m, \alpha_m, \beta_m), \nonumber
    \label{re:NIG_sum}
\end{equation}
where $M$ denotes the number of interactions. Although simple in form, unfortunately, it is intractable to infer the parameters for the fused NIG distribution since there is no closed-form solution. Therefore, inspired by multi-modal learning~\cite{re:TMR} and multi-source learning~\cite{re:BBLR}, we employ the simple NIG summation operation to approximately solve this problem
\begin{equation}
\begin{split}
    NIG(\gamma, \omega, \alpha, \beta) & \triangleq NIG(\gamma_1, \omega_1, \alpha_1, \beta_1) \\ 
    &\oplus NIG(\gamma_2, \omega_2, \alpha_2, \beta_2) \\
    &\oplus \cdots  \\
    &\oplus NIG(\gamma_M, \omega_M, \alpha_M, \beta_M),
\end{split}
\label{eq:fusionNIG}
\end{equation}
where $M$ denotes the number of interaction, $\oplus$ denotes the summation operation of two NIG distributions as follows,
\begin{equation}
		\oplus\left\{
		\begin{aligned}
			\gamma&=(\omega_1+\omega_2)^{-1}(\omega_1\gamma_1+\omega_2\gamma_2), \\
			\omega & =  \omega_1 + \omega_2, \\
			\alpha & =  \alpha_1+\alpha_2+\frac{1}{2}, \\
            \beta & = \beta_1+\beta_2+\frac{1}{2}\omega_1(\gamma_1-\gamma)^2 + \frac{1}{2}\omega_2(\gamma_2-\gamma)^2. \nonumber
		\end{aligned}
		\right.
\end{equation}
The NIG summation can reasonably make use of predictions with different qualities. Specifically, the parameter $\omega$ indicates the confidence of a NIG distribution for the mean $\gamma$. If one matte is more confident with its prediction, then it will contribute more to the final prediction. Moreover, $\beta$ directly reflects both aleatoric uncertainty and epistemic uncertainty which consists of two parts, i.e., the sum of $\beta_1$ and $\beta_2$ from multiple mattes and the variance between the final prediction and that of every single matte. 


\subsection{Aleatoric Uncertainty-based Refinement}
\label{sec:alea}

Some methods \cite{re:reduceAl, re:aleaAug} constrain invariant predictions for the simulated inherent noise by data augmentation, i.e., enhancing the robustness by explicitly modeling noise to reduce the aleatoric uncertainty. Given the noise $\epsilon \sim \mathcal N(0, \varepsilon)$ for input $\chi$, a simple way to reduce aleatoric uncertainty is constraining consistent prediction for samples from $\mathcal N(\chi, \varepsilon)$~\cite{re:reduceAl}. However, characterizing the noise of the data requires additional self-supervised training, such as image reconstruction. To simplify the training steps, we propose a plug-and-play module $\mathcal R_{\phi}$ to reduce aleatoric uncertainty and also to refine the matting details. Instead of modeling the noise during the input, we directly model the output noise in terms of the aleatoric uncertainty $\mathbb E(\sigma^2)$. In other words, we regard the matting output $\gamma$ as $\chi$, and the noise $\epsilon \sim \mathcal N(0, \mathbb E(\sigma^2))$, and then, we attempt to keep the consistent prediction for data sampling from $\mathcal N(\gamma, \mathbb E(\sigma^2))$. 
Furthermore, we use the variance $Var(\sigma^2)$ to filter out the regions whose aleatoric uncertainty $\mathbb E(\sigma^2)$ may be inaccurate. The $Var(\sigma^2)$ \cite{re:in-Gamma} is defined as
\begin{equation}
    Var[\sigma^2] = \frac{\beta^2}{(\alpha-1)^2(\alpha-2)}, \nonumber
\end{equation}
where $\alpha > 2$.

The objective of the refinement module is to restore high-aleatoric uncertainty matting details without redundant calculation, thus it only concentrates on local patch refinement. We first obtain the coarse matte $\gamma_s$, sampling once from $\mathcal N(\gamma, \mathbb E(\sigma^2))$ due to small variance as shown in \cref{fig:framework}. Then, we use OTSU \cite{re:OTSU} and $Var[\sigma^2]$ to adaptively select the pixels of reliable high aleatoric uncertainty. Finally, the $32 \times 32$ patches centered on the selected pixels corresponding to $\gamma_s$ are fed into our refinement module $\mathcal R_{\phi}$, and the obtained predictions replace the coarse matte corresponding position to obtain the refined alpha matte $\gamma_r$.

\begin{table*}[t]
\centering
\caption{Comparison results on the benchmarks P3M-500-P~\cite{re:gfm} and P3M-500-NP~\cite{re:gfm}. $\ddagger,\dagger$ denote predictions without and with user map, respectively. For all metrics, the smaller value indicates the better performance.}
\label{tab:P3M}
\resizebox{\linewidth}{!}{
\begin{tabular}{l|c|c|c|c|c|c|c||c|c|c|c|c|c|c}
\hline
& \multicolumn{7}{c||}{P3M-500-P} & \multicolumn{7}{c}{P3M-500-NP} \\\hline
Method&  SAD&  MSE&  MAD&  Grad & SAD$_{bf}$ & SAD$_t$ & Conn & SAD&  MSE&  MAD&  Grad & SAD$_{bf}$ & SAD$_t$ & Conn \\ \hline
SHM \cite{re:SHM}&  26.84&  1.26&  1.65&  20.18 &  16.90& 9.94& 23.30 &  30.20&  1.46&   1.93&  20.31& 17.99& 12.21& 26.06 \\
U$^2$Net \cite{re:u2net}&  73.48&  1.99&  4.51&  33.06&  48.54&  26.91& 53.81  &  70.67&  1.89&  4.51& 34.89& 42.75&  27.91& 53.29 \\
MODNet \cite{re:modnet}&  23.86& 1.11& 1.46&  23.74 &  16.40&  7.46&  21.02 &  25.39&  1.20&  1.61&  21.15& 17.41& 7.98&  22.22\\
GFM \cite{re:gfm}&  12.90&  0.58&  0.79&  14.61 & 5.98&  6.93& 11.33 &  17.01& 0.85&  1.09&  14.54&  8.84&  8.17 &14.86\\
P3MNet \cite{re:P3M}&  12.73&  0.56&  0.78&  13.89&  5.95&  6.78 &11.14 & 16.49&  0.80&  1.05&  12.75&  8.97 & 7.54 & 14.35 \\ \hline

SHM (dugMatting) $\ddagger$ &  21.43&  1.26&  1.51&  17.82&  11.57& 10.07& 19.32 & 39.67&  1.66&  2.43&  17.23&  28.27& 11.40 &33.88\\
U$^2$Net (dugMatting) $\ddagger$&  60.21&  1.76&  4.24&  28.74&  31.66& 28.55& 47.34 & 82.67&  2.29&  5.07& 31.65& 51.29&  31.38 &60.12\\
MODNet (dugMatting) $\ddagger$&  18.15& 0.72& 1.04&  15.57 &  9.59&  8.55 & 16.75& 35.66 & 1.49 & 2.07 & 16.04 &24.26 & 11.40 & 32.83\\
GFM (dugMatting) $\ddagger$&  9.25&  0.40&  0.63&  13.79 &  3.18& 6.71 & 9.29 & 19.01&  0.86&  1.14&  14.14&  9.27& 9.73 & 16.45\\
P3MNet (dugMatting) $\ddagger$&  10.08&  0.46&  0.69&  14.61&  4.03& 7.01 & 10.30 & 16.12&  0.66&  0.94&  14.15& 6.92 & 9.18 & 13.81\\
$\triangle$ Average gain $\ddagger$& -2.13 & -0.17 & -0.21 & 0.13 & -6.01 & 1.86 & -1.54 & 14.31 & 0.15 & 0.29 & 1.02 & 6.67&5.02&7.86\\ \hline


SHM (dugMatting) $\dagger$& 13.87&  0.36&  0.85&  15.21&  4.83& 9.04 & 11.42 & 18.22&  0.51&  1.11&  13.99&  6.57 & 11.65 & 15.16\\
U$^2$Net (dugMatting) $\dagger$&  35.23&  1.35&  2.16&  19.32&  22.78 & 12.45 & 32.76 & 39.86&  1.67&  2.44& 20.12& 21.26&  18.60 & 33.91\\
MODNet (dugMatting) $\dagger$&  9.62& 0.29& 0.55&  12.88 &  2.63&  6.98 & 9.05& 11.08&  0.33&  0.64&  11.75& 3.19&  7.88 & 10.99\\
GFM (dugMatting) $\dagger$&  7.90&  0.23&  0.46&  12.31&  1.29& 6.60 & 6.59 & 9.55&  0.28&  0.55&  11.01&  2.12& 7.42 & 7.95\\
P3MNet (dugMatting) $\dagger$&  7.72&  0.22&  0.45&  12.56&  1.01& 6.71& 6.42& 8.79&  0.24&  0.51&  11.08&  1.34& 7.47 &7.23\\
$\triangle$ Average gain $\dagger$&  \textbf{-15.09} & \textbf{-0.61} & \textbf{-0.94} & \textbf{-6.64} & \textbf{-12.24} & \textbf{-3.24} & \textbf{-10.87} & \textbf{-14.45} & \textbf{-0.63} & \textbf{-0.98} & \textbf{-7.13} & \textbf{-12.29} & \textbf{-2.15} & \textbf{-11.10}\\ \hline

\end{tabular}
}
\end{table*}

\begin{table*}[t]
\centering
\caption{Quantitative comparison results of natural matting on Composition-1K~\cite{re:DIM} benchmark.}
\label{tab:DIM}
\resizebox{\linewidth}{!}{
\begin{tabular}{l|c|c|c|c|c|c}
\hline
Method& User Map & SAD $(10^{3})\downarrow$ & MAD$\downarrow$ & MSE $(10^{-3})\downarrow$ & \quad Grad $\downarrow$ \quad&  \quad Conn$\downarrow$ \quad \\ \hline
Learning Based Matting \cite{re:LBDM} & Trimap &  113.9 & 0.0501 &  48.0&  91.6&  122.2\\
Closed-Form Matting \cite{re:ACFS} & Trimap&  168.1& 0.0739 &  91.0&  126.9&  167.9\\
KNN Matting \cite{re:KM} & Trimap &  175.4& 0.0771& 103.0& 124.1&  176.4 \\
Deep Image Matting \cite{re:DIM} & Trimap&  50.4& 0.0221&  14.0&  31.0&  50.8\\
AlphaGan \cite{re:alphaM} & Trimap &  52.4& 0.0231 &  30.0&  38.0&  - \\ 
IndexNet \cite{re:indexM} & Trimap&  45.8& 0.0201 &  13.0&  25.9&  43.7\\
HAttMatting \cite{re:HattM} & Trimap&  44.0& 0.0193&  7.0&  29.3&  46.4\\ 
AdaMatting \cite{re:AdaM}& Trimap&  41.7& 0.0183&  10.0&  16.8&  -\\
sampleNet \cite{re:SampleNet}& Trimap&  40.4& 0.0177&  9.9&  -&  -\\
Fine-Grained Matting \cite{re:FGM}& Trimap&  37.6& 0.0165&  9.0&  18.3&  35.4\\
Context-Aware Matting \cite{re:CAIM}& Trimap&  35.8& 0.0157 &  8.2&  17.3&  33.2\\
GCA Matting \cite{re:GCAM} & Trimap&  35.3& 0.0155&  9.1&  16.9&  32.5\\
HDMatt \cite{re:HDMatting} & Trimap&  33.5& 0.0147&  7.3&  14.5&  29.9\\
MG Matting \cite{re:MGMatting} & Mask&  31.5&  0.0138&  6.8&  13.5&  27.3\\
TIMNet \cite{re:TIMINet}& Trimap&  29.1& 0.0128&  6.0&  11.5&  25.4\\
SIM \cite{re:SIM}& Mask&  28.0& 0.0123&  5.8&  10.8&  24.8\\
MatteFormer \cite{re:matteFormer} & Trimap&  23.8&  0.0104&  4.0&  8.7&  18.9\\ \hline\hline
MG Matting (dugMatting) & w/o & 36.5 & 0.0161 & 8.5 & 17.8 & 33.6 \\
MG Matting (dugMatting) & 1-Selection& 32.3 & 0.0142 & 7.1 & 14.2 & 28.6 \\
MG Matting (dugMatting) & 2-Selection& 30.2 & 0.0132& 6.4 & 11.8 & 26.1 \\ \hline
MatteFormer (dugMatting) & w/o & 34.1 & 0.0149 & 5.7 & 15.6 & 31.2 \\
MatteFormer (dugMatting) & 1-Selection& 25.8 & 0.0112 & 4.3 & 9.7 & 22.3 \\
MatteFormer (dugMatting) & 2-Selection& 23.4 & 0.0102& 3.9 & 7.2 & 18.8 \\ \hline
\end{tabular}
}
\end{table*}

\subsection{Optimization}
We train the proposed dugMatting in two stages to enhance the stability, i.e., optimizing  $\mathcal F_{\theta}$ to empower the epistemic uncertainty-based interaction and $\mathcal R_{\phi}$ to refine details.


For the optimization of $\mathcal F_{\theta}$, intuitively, we need to simulate and supervise different predictions in real interaction process, including the initial prediction, the prediction after interaction, and the fused prediction. To simplify the training process, we analyze the purpose of supervision in the three predictions. The supervision of the initial prediction aims to train the network to conduct matting without user map. The supervision of the prediction after interaction aims to relate the network predictions to interaction, which assumes the user map is generated according to epistemic uncertainty. The supervision of the fused prediction aims to stabilize the the fusion result. Based on above analysis, we can jointly supervise the initial prediction and the prediction after interaction by generating random user map $U$ including the empty case. The details of user map can be found in \cref{appendix:UM}. The supervision of the fused prediction can be removed because the fusion strategy in \cref{eq:fusionNIG} is exactly for stabilization. Therefore, the simplified supervision is similar to the previous interactive matting methods \cite{re:improveM}, which only needs to pass through the model once in each iteration. The loss of the first stage can be expressed as
 \begin{equation}
    \mathcal L_{stage1}= \mathcal L_{NIG}(\gamma, \omega, \alpha, \beta;\theta) + \mathcal L_{M}(\gamma;\theta_\gamma), \nonumber
    \label{eq:unMatting}
\end{equation}
where minimizing $\mathcal L_{NIG}$ optimizes the parameters of NIG distribution to replace the regression loss (e.g., $l_1$ loss or $l_2$ loss) in common matting methods, and $\mathcal L_{M}$ denotes the additional terms (e.g., Laplacian loss \cite{re:gfm}) about matte in the original matting methods.

For the optimization of $\mathcal R_{\phi}$, we first freeze the parameters $\theta$ of $\mathcal F_\theta$. Then, we can obtain the $\gamma_s$ and $k$ patches of interest $\{\gamma_s^p\}^k$ according to \cref{sec:alea}. The $l_1$ distance with the ground truth $y$ in matte and gradient map are used for supervision. The loss of the second stage is
\begin{equation}
    \mathcal L_{stage2}=||y- \gamma_r||_1+||\triangledown y - \triangledown \gamma_r||_1, \nonumber
\end{equation}
where $\gamma_r =\mathcal R_\phi(\gamma_s^k, \gamma_s), \gamma_s=(\gamma + \epsilon), \epsilon \sim \mathcal N(0, \mathbb E(\sigma^2))$.

\section{Experiments}
\subsection{Experimental Setup}
\textbf{Dataset.} We conduct extensive experiments on standard natural matting dataset Composition-1k \cite{re:DIM} and the real-world portrait dataset P3M-10K \cite{re:P3M}. Composition-1k \cite{re:DIM} contains 43,100 synthetic images for training and 1000 synthetic images for testing. P3M-10K \cite{re:P3M} consists of $10,000$ anonymized high-resolution portrait images with face obfuscation, containing $9,421$ images for training and $500$ images denoted as P3M-500-P for testing. Besides, for P3M-10K there are additional $500$ public Internet images without face obfuscation to test the matting performance on regular portrait images, denoted as P3M-500-NP.

\textbf{Implementation Details}. For class-specific matting, we train all models with the same data augmentations setting for a fair comparison, including random horizontal flipping, random blurring, random sharpen, random shadow, and then random cropping to $512 \times 512$ in the end. All models are optimized using the Adam optimizer~\cite{re:Adam_optimizer}, and the base learning rate is set to $1 \times 10 ^{-3}$ with the cosine learning rate scheduler~\cite{re:cosine}, 100 epochs iteration, and batch size of 16. For natural image matting, we use the standard setting as specified by MatteFormer \cite{re:matteFormer}.
Our implementation \footnote{Code is available at \url{https://github.com/Fire-friend/dugMatting}.} is based on the open source framework Pytorch . All the experiments were run on two GeForce RTX 3090 GPUs. 

\textbf{Evaluation Metrics}. For Composition-1k, we employ multiple quantitative metrics, i.e., sum of absolute differences (SAD), mean absolute difference (MAD), mean squared error (MSE), gradient (Grad), and connectivity (Conn). For P3M-10K, we also adopt the above metrics and report the additional SAD$_{bf}$ and SAD$_t$ to compute the SAD within the foreground-background regions and transition regions.

\subsection{Quantitative Analysis}
\textbf{Class-specific Matting.} To validate our methods on class-specific matting task, we compare our algorithm with state-of-the-art trimap-free methods~\cite{re:SHM, re:u2net, re:modnet, re:gfm,re:P3M} on real-world portrait dataset~\cite{re:P3M}. As shown in \cref{tab:P3M}, dugMatting without interaction outperforms the original trimap-free methods on P3M-500-P, demonstrating that the way of modeling uncertainty can improve the matting performance. In addition, dugMatting significantly improves performance when introducing once interaction, particularly by roughly 50\% on P3M-500-NP, demonstrating that the interaction is still useful even when dealing with data from different domains.

\textbf{Natural Image Matting.} The natural image matting expects to extract the interesting foreground with the guidance of user interaction. We first investigate the natural matting methods \cite{re:LBDM,re:KM,re:DIM, re:ACFS,re:alphaM,re:indexM,re:HattM,re:AdaM,re:SampleNet,re:FGM,re:CAIM,re:GCAM,re:HDMatting, re:MGMatting} on Composition-1k \cite{re:DIM}. Then, we employ the effective MG Matting \cite{re:MGMatting} and MatteFormer \cite{re:matteFormer} as foundation models, integrating our method to validate the performance on natural matting task. Since the Composition-1k is a synthetic set, it allows for the extraction of target objects without any initial interaction. However, when dealing with arbitrary images in real-world, we suggest providing an initial user map through a single click and then utilizing our method for further interaction. The quantitative results are shown in \cref{tab:DIM}. With only one or two interactions, our dugMatting outperforms advanced trimap-based matting algorithms. The reason is that reducing the decomposed uncertainties can accurately improve the matte. We also conduct experiments to compare the efficiency of existing interaction methods in \cref{appendix:RC}.

\subsection{Qualitative Analysis}

\begin{figure*}[htb]
	\begin{center}
    \captionsetup[subfigure]{labelsep=none,format=plain,labelformat=empty}
    \subfloat[Input]{\includegraphics[width=0.15\linewidth]{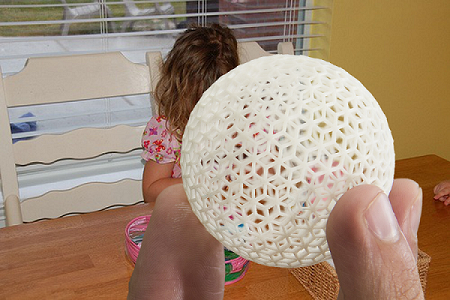}}\hskip 3pt
    \subfloat[Trimap]{\includegraphics[width=0.15\linewidth]{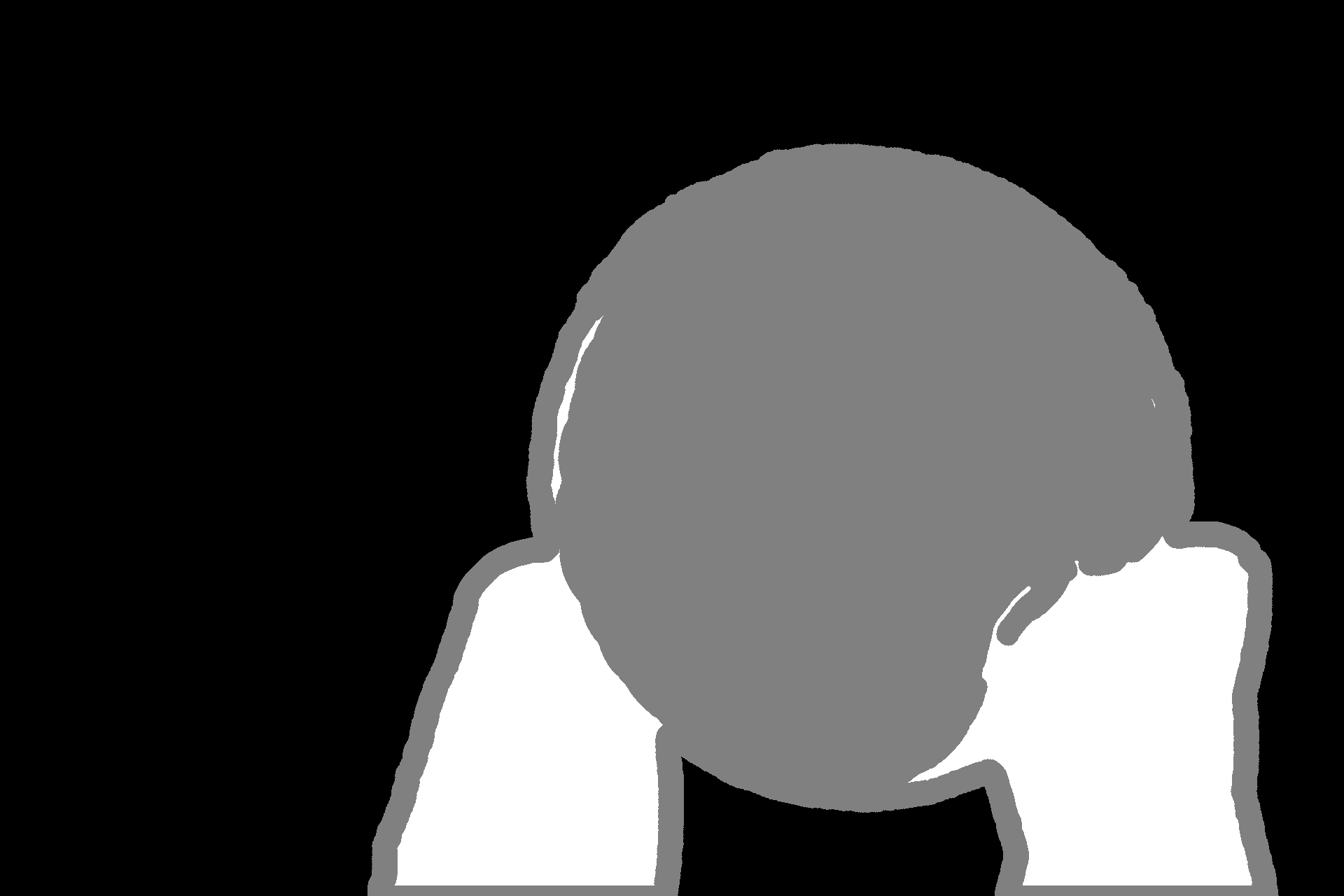}}\hskip 3pt
    \subfloat[GT]{\includegraphics[width=0.15\linewidth]{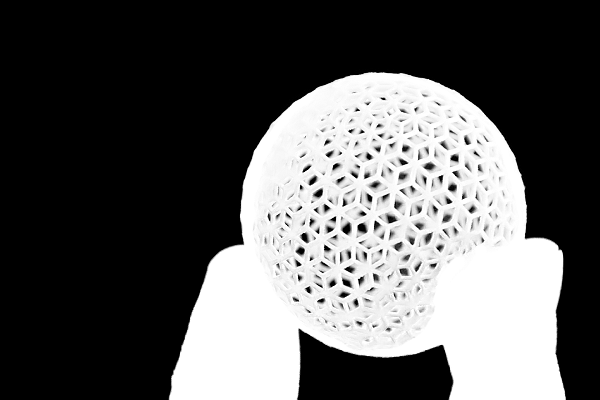}}\hskip 3pt
    \subfloat[Closed-Form]{\includegraphics[width=0.15\linewidth]{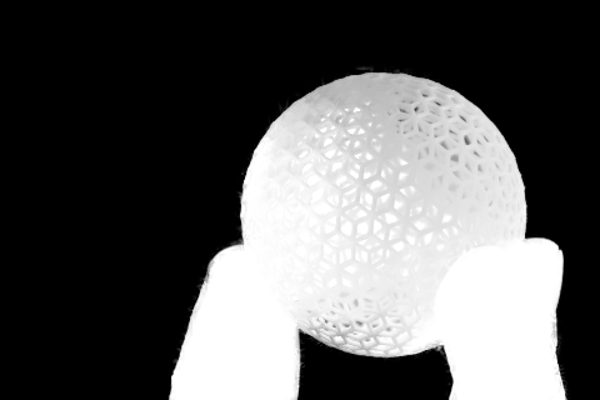}}\hskip 3pt
    \subfloat[Learning]{\includegraphics[width=0.15\linewidth]{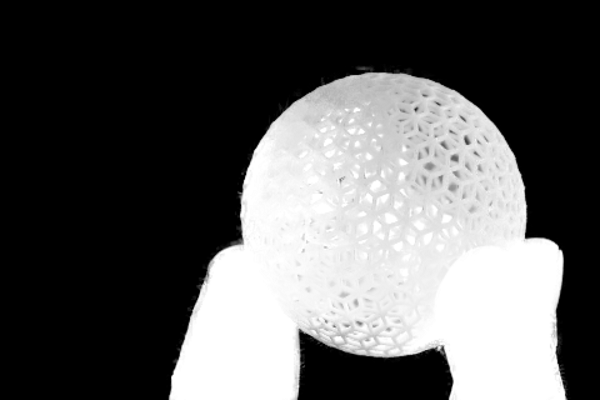}}\hskip 3pt
    \subfloat[DIM]{\includegraphics[width=0.15\linewidth]{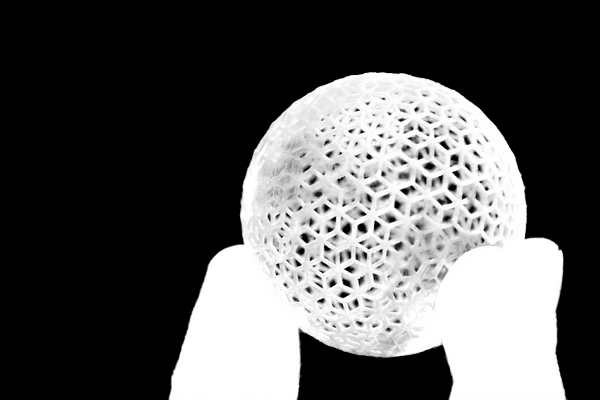}}\vskip -0.1in
    \subfloat[IndexNet]{\includegraphics[width=0.15\linewidth]{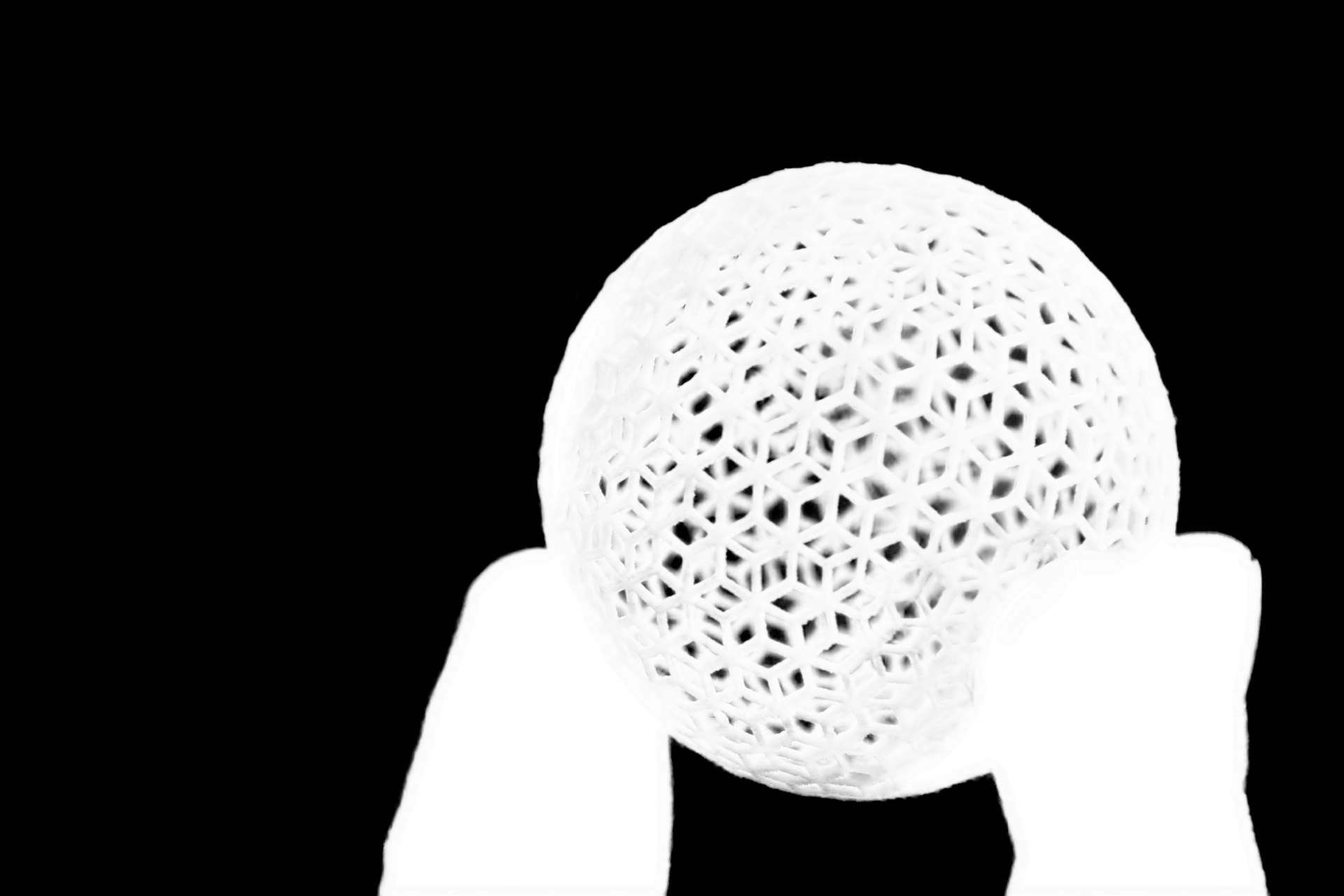}}\hskip 3pt
    \subfloat[CAM]{\includegraphics[width=0.15\linewidth]{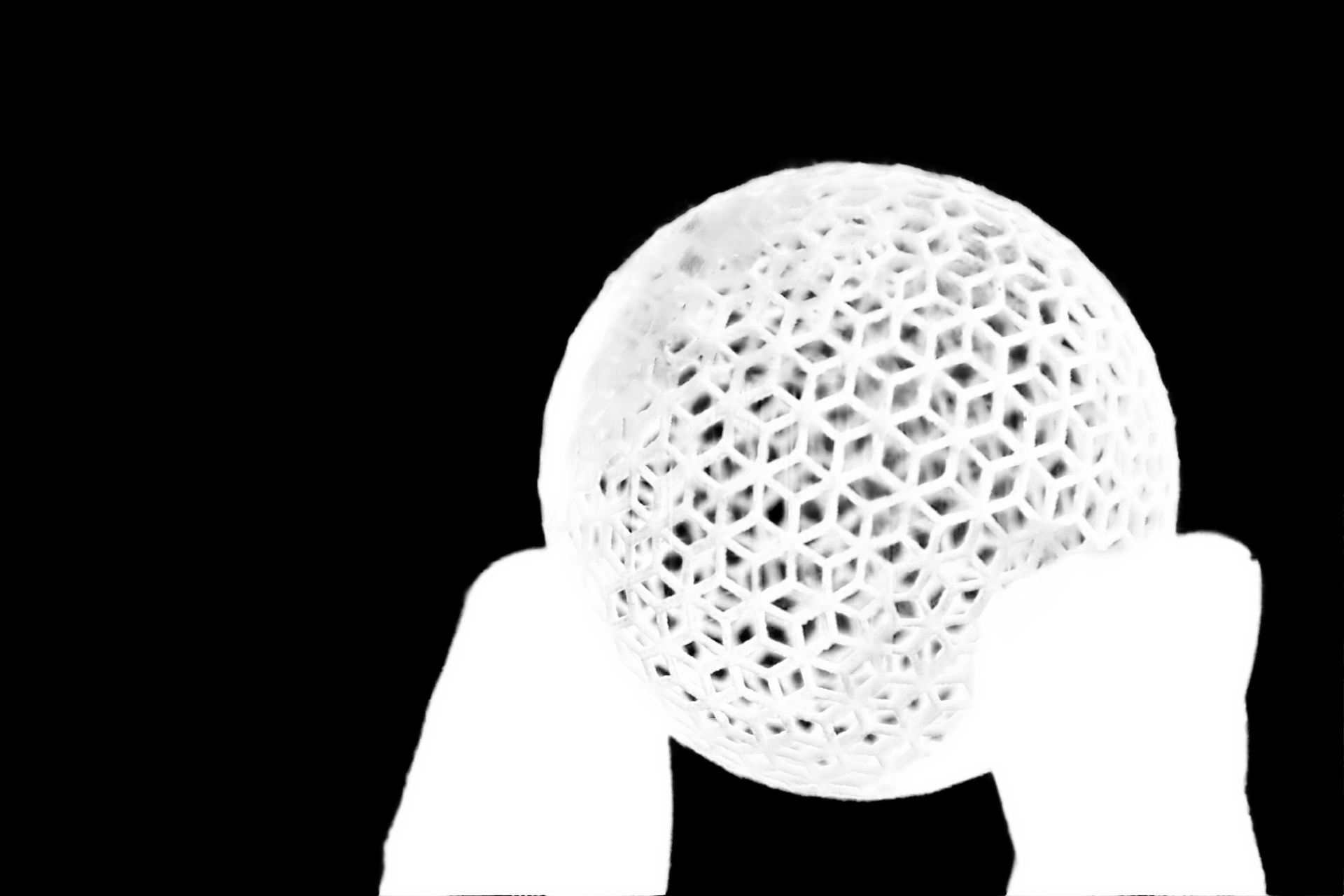}}\hskip 3pt
    \subfloat[GCA Matting]{\includegraphics[width=0.15\linewidth]{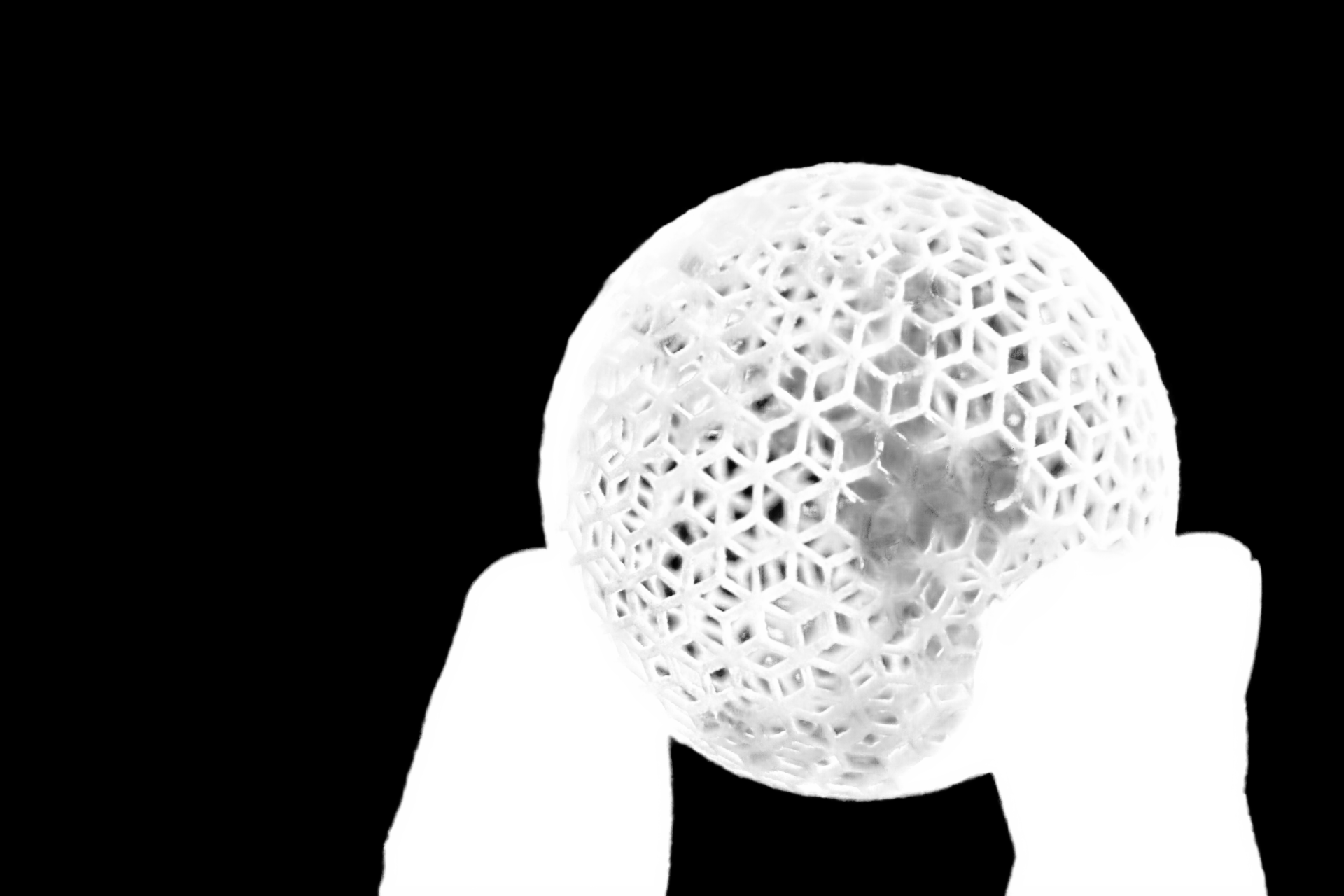}}\hskip 3pt
    \subfloat[MG Matting]{\includegraphics[width=0.15\linewidth]{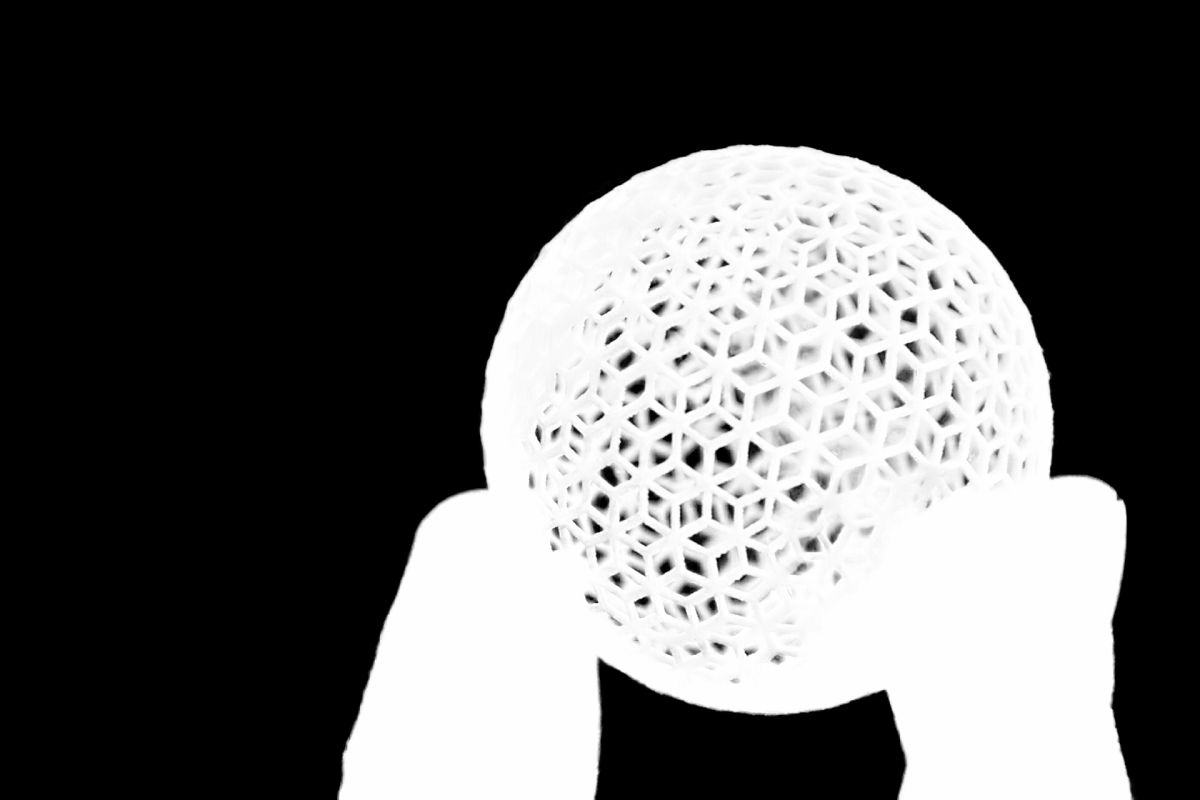}}\hskip 3pt
    \subfloat[MatteFormer]{\includegraphics[width=0.15\linewidth]{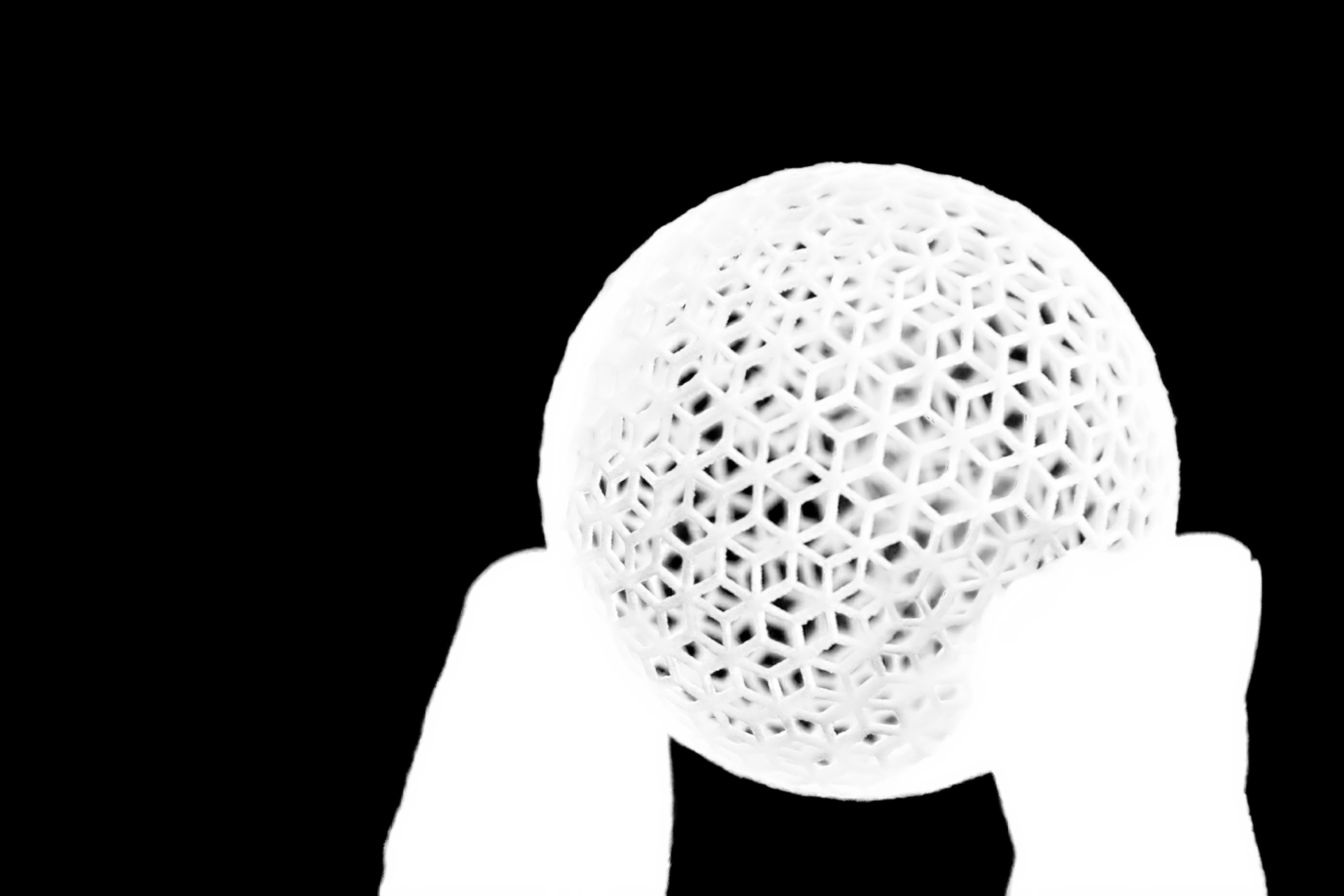}}\hskip 3pt
    \subfloat[Ours]{\includegraphics[width=0.15\linewidth]{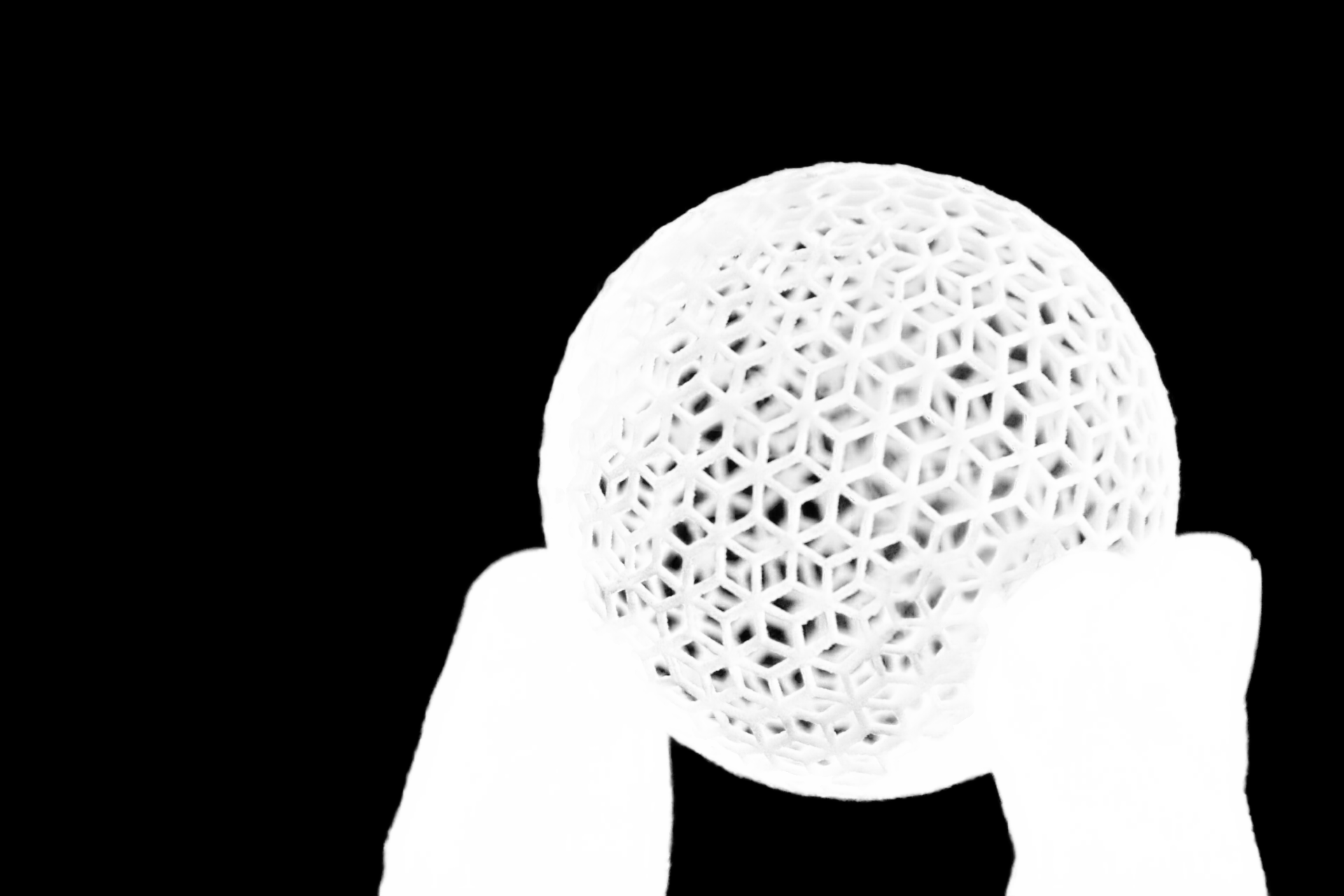}}\vskip -0.1in
    \subfloat[Input]{\includegraphics[width=0.15\linewidth]{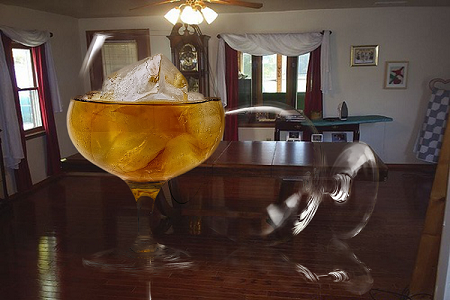}}\hskip 3pt
    \subfloat[Trimap]{\includegraphics[width=0.15\linewidth]{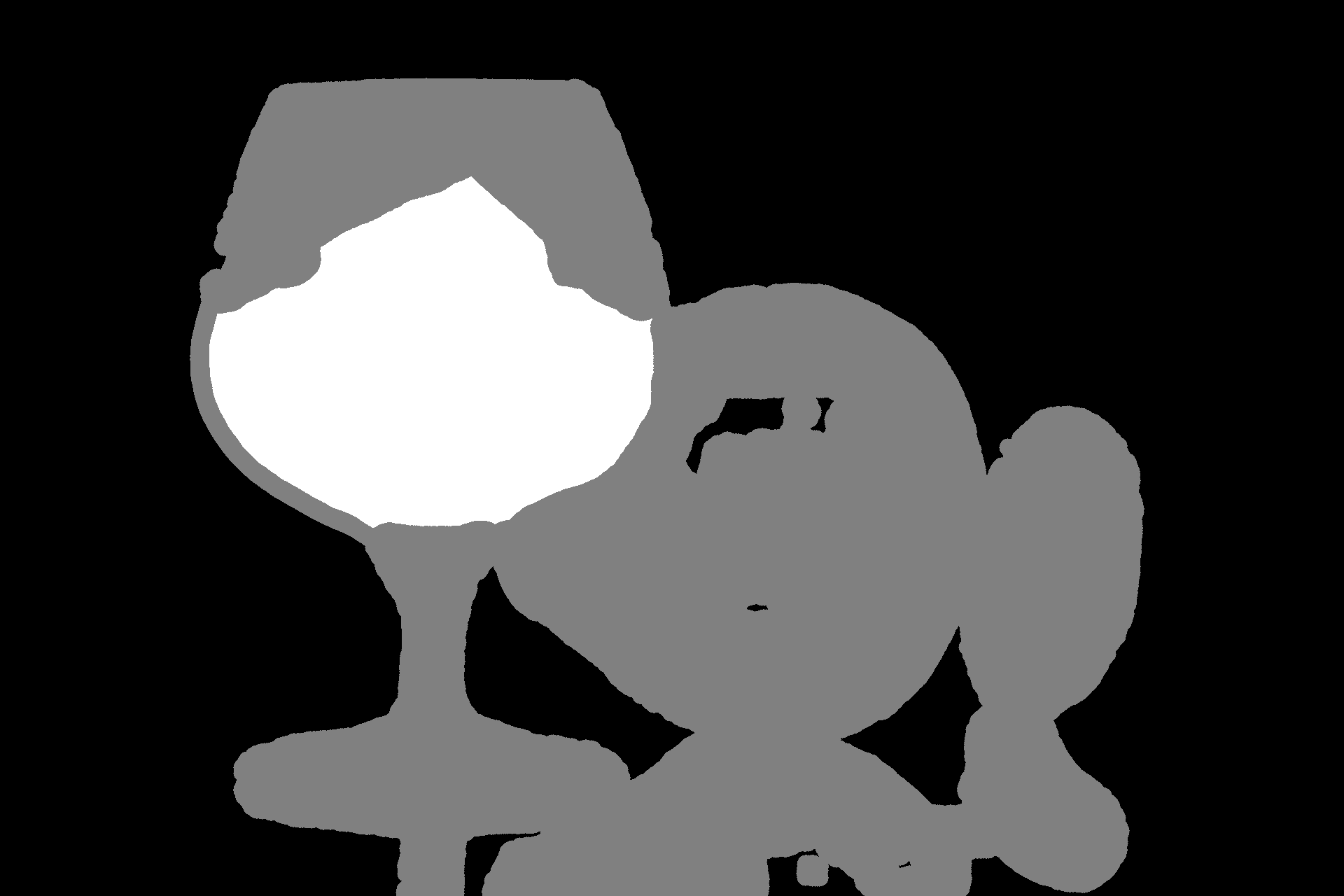}}\hskip 3pt
    \subfloat[GT]{\includegraphics[width=0.15\linewidth]{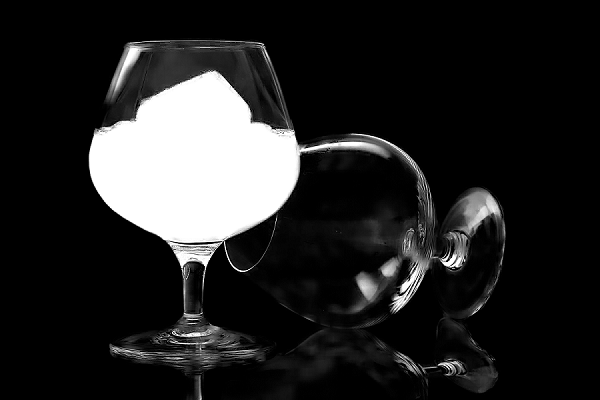}}\hskip 3pt
    \subfloat[Closed-Form]{\includegraphics[width=0.15\linewidth]{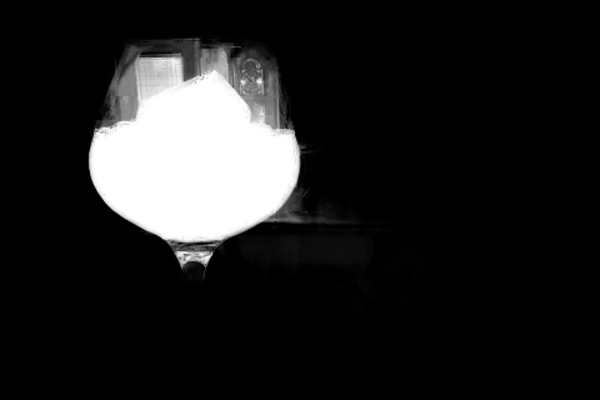}}\hskip 3pt
    \subfloat[Learning]{\includegraphics[width=0.15\linewidth]{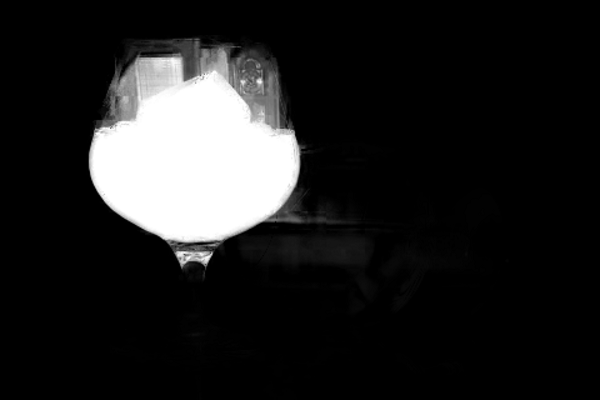}}\hskip 3pt
    \subfloat[DIM]{\includegraphics[width=0.15\linewidth]{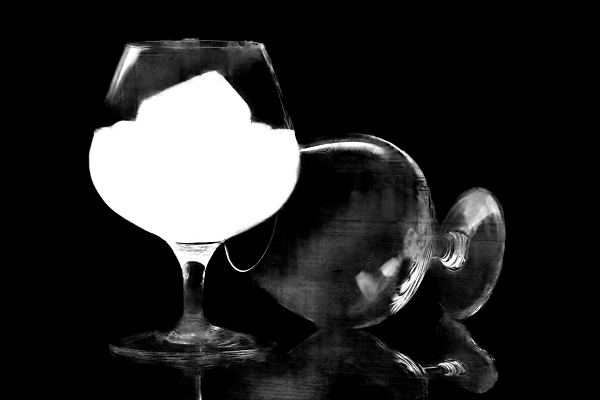}}\vskip -0.1in
    \subfloat[IndexNet]{\includegraphics[width=0.15\linewidth]{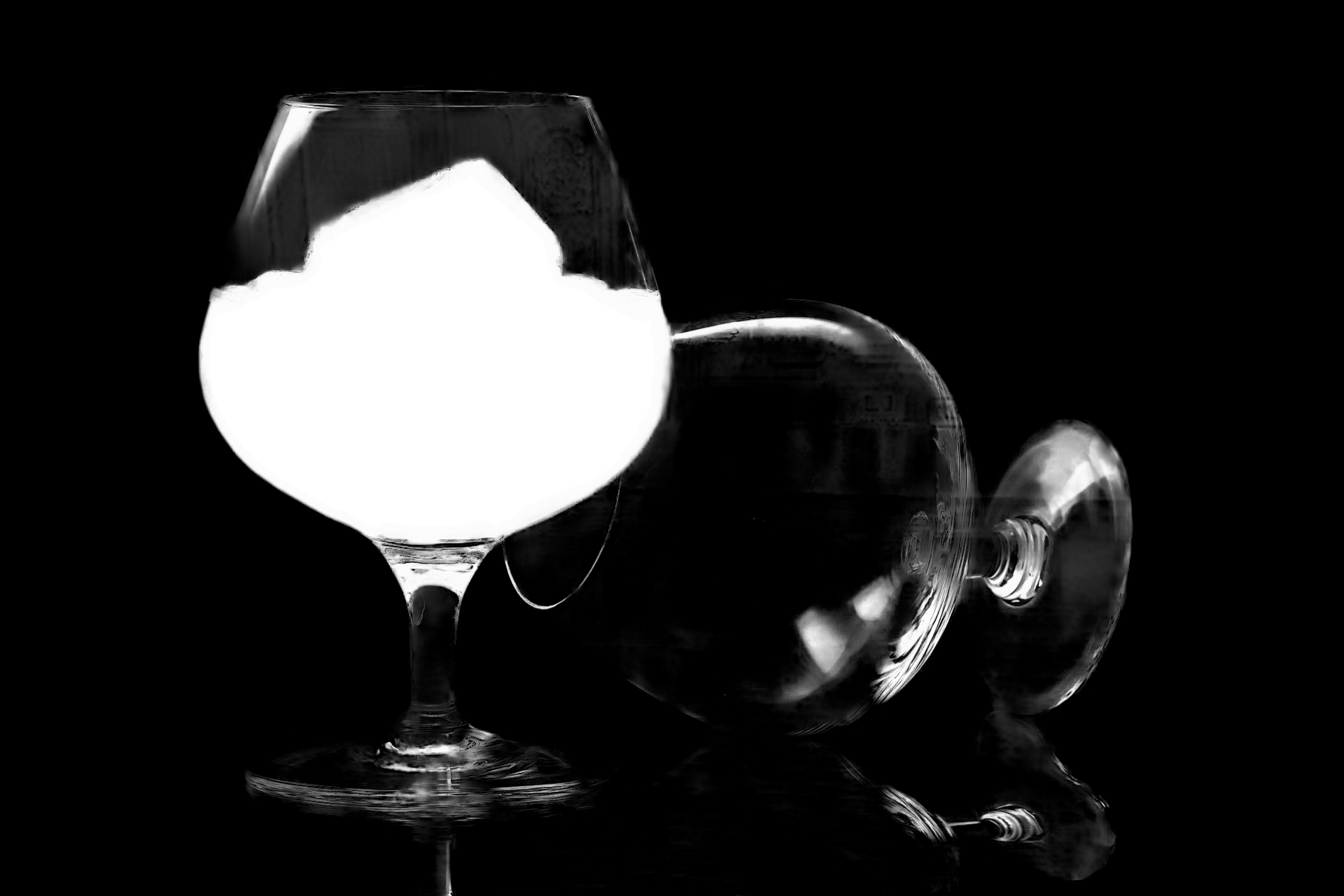}}\hskip 3pt
    \subfloat[CAM]{\includegraphics[width=0.15\linewidth]{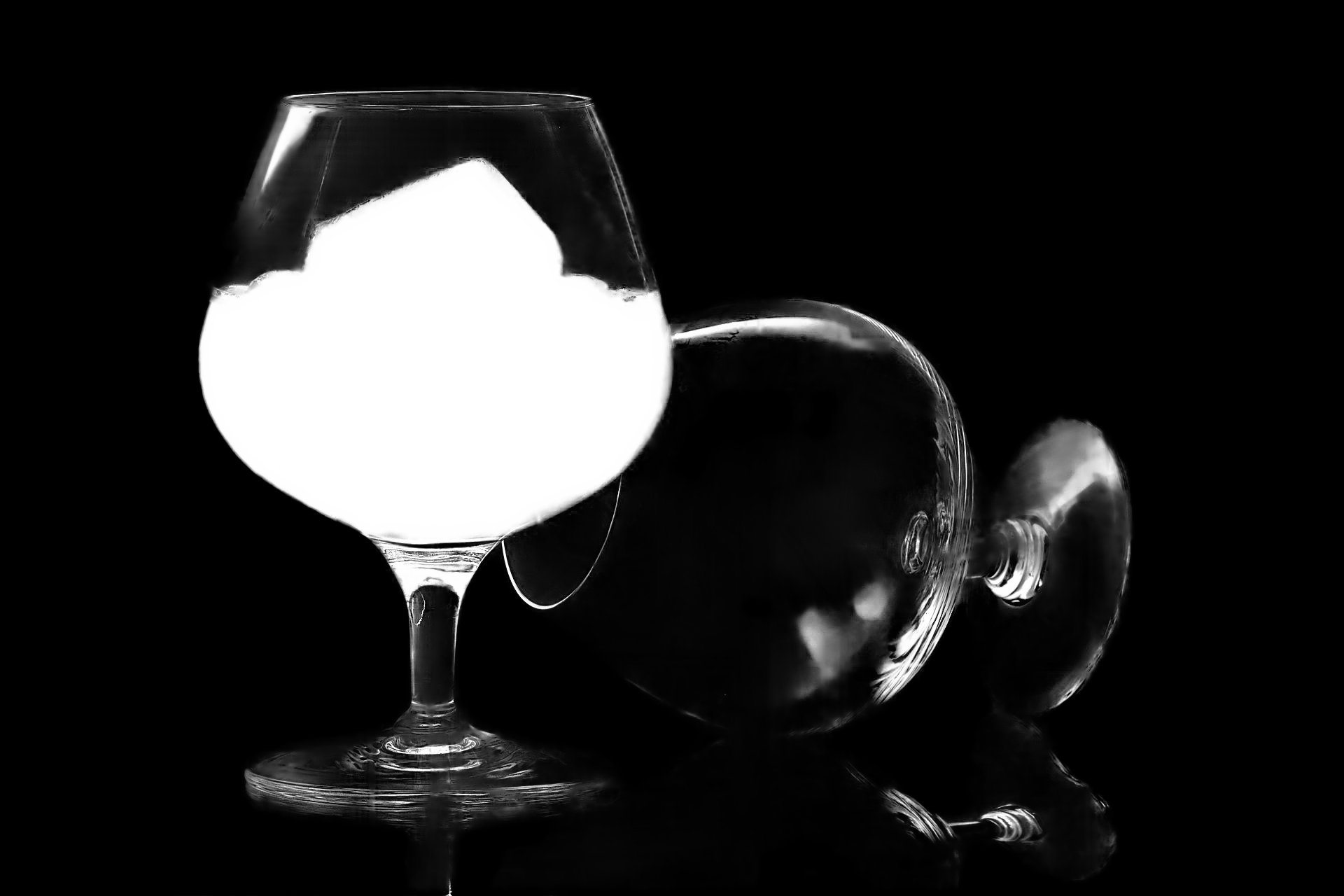}}\hskip 3pt
    \subfloat[GCA Matting]{\includegraphics[width=0.15\linewidth]{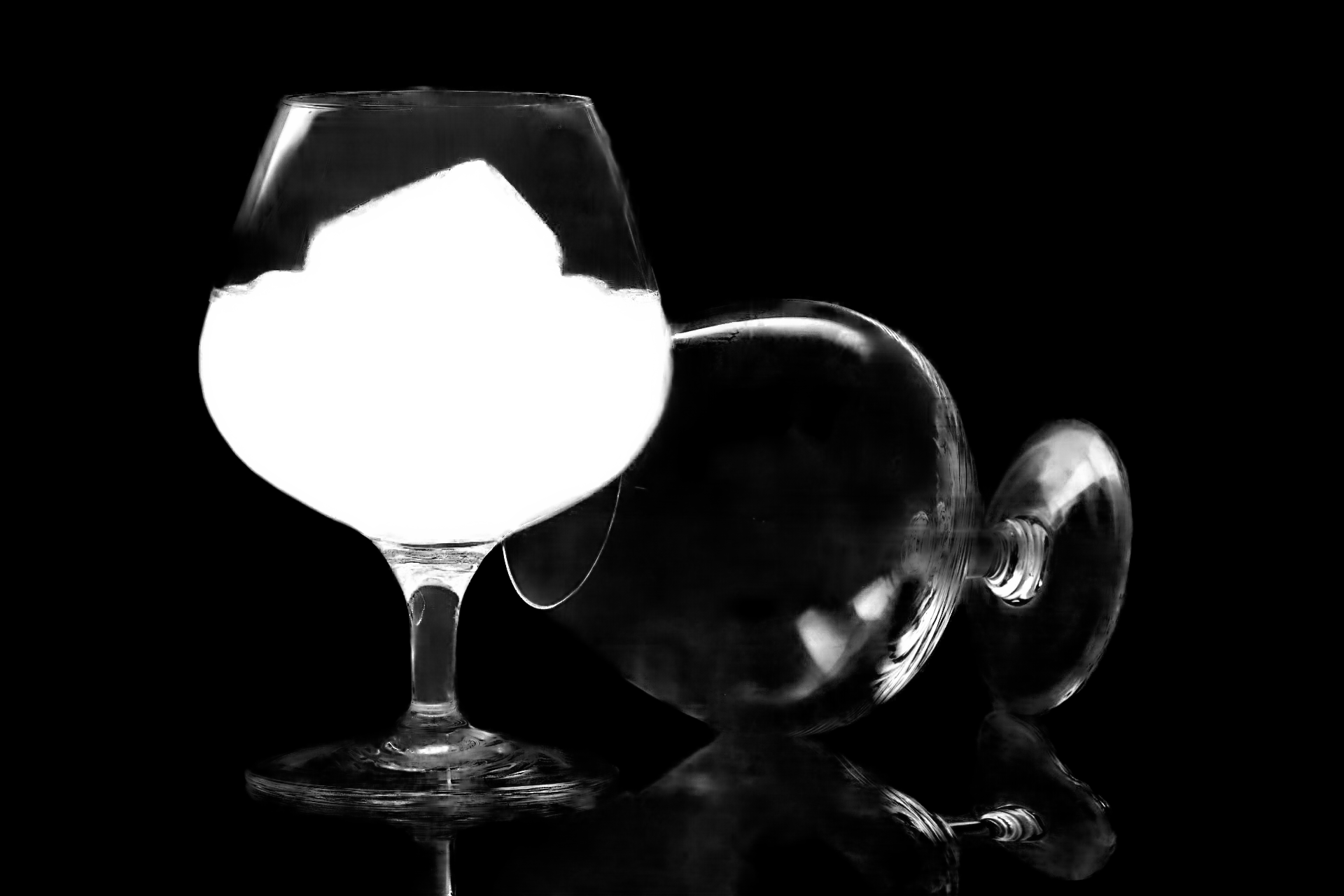}}\hskip 3pt
    \subfloat[MG Matting]{\includegraphics[width=0.15\linewidth]{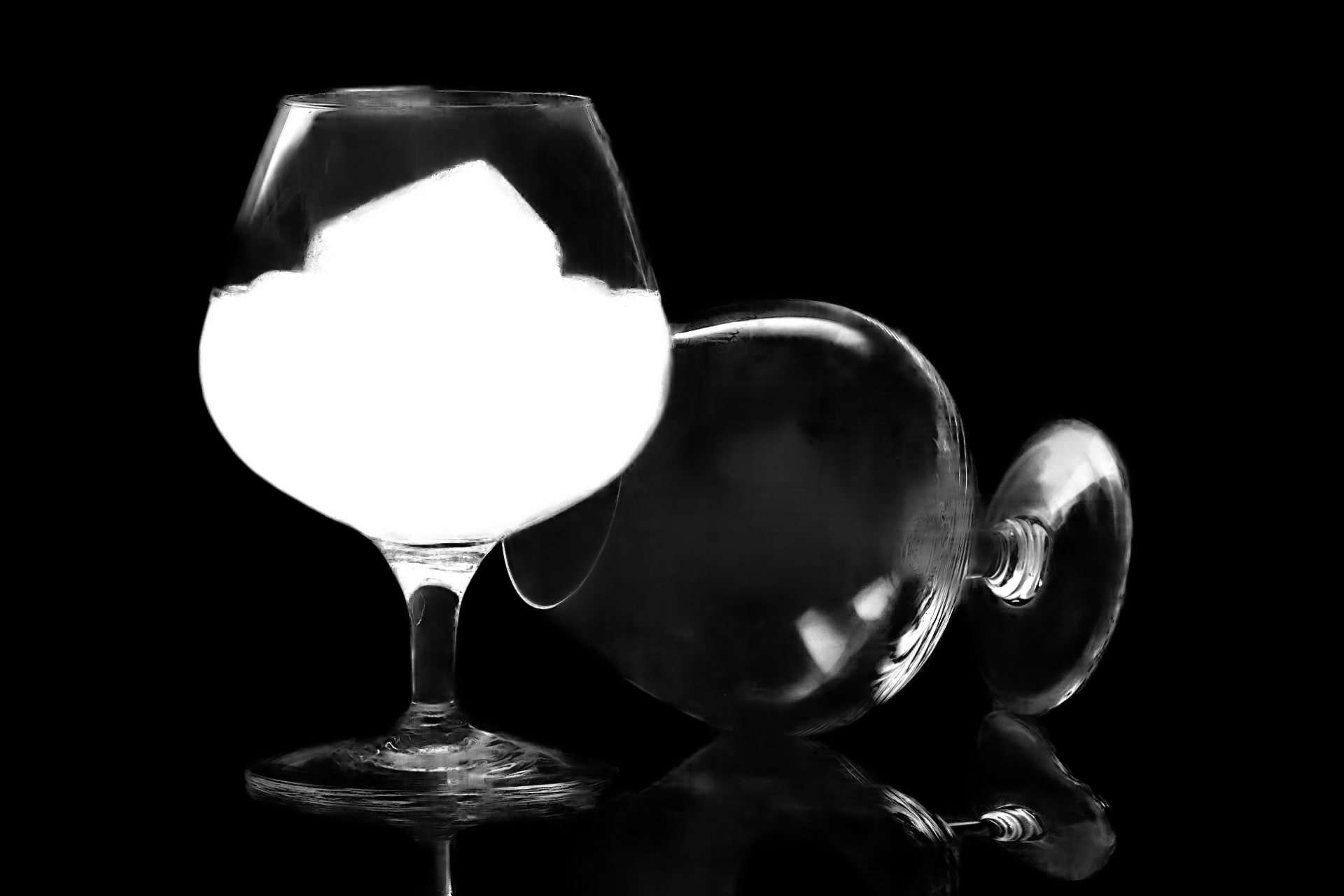}}\hskip 3pt
    \subfloat[MatteFormer]{\includegraphics[width=0.15\linewidth]{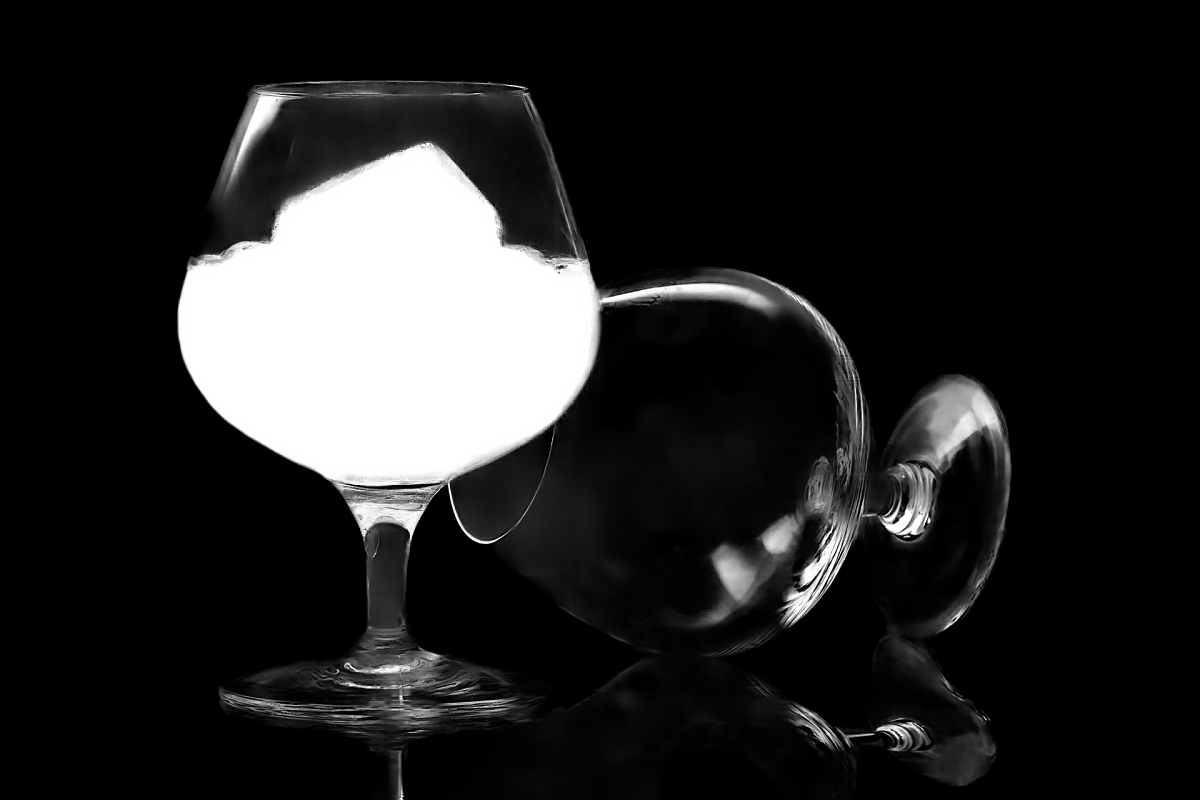}}\hskip 3pt
    \subfloat[Ours]{\includegraphics[width=0.15\linewidth]{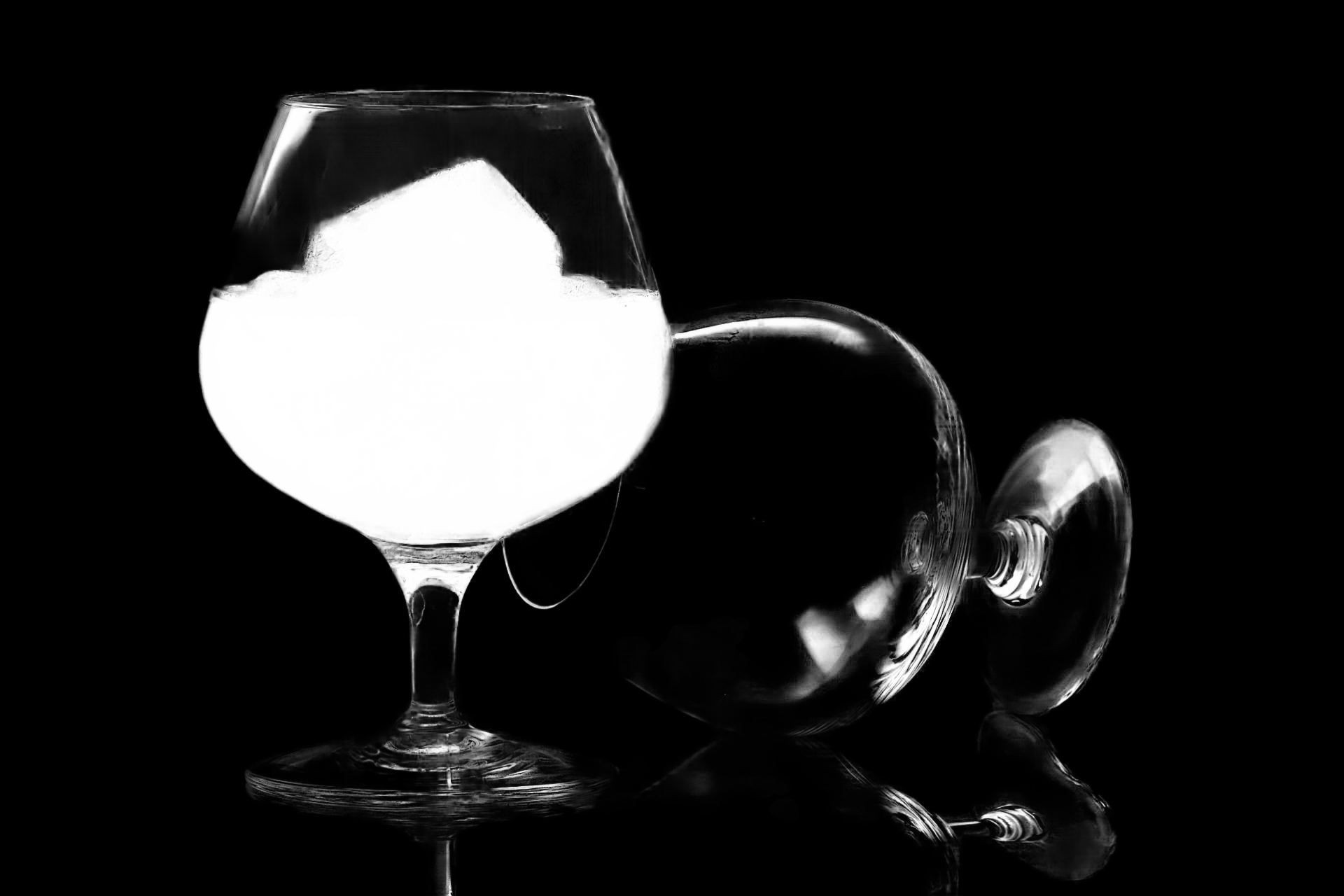}}
    \end{center}
	\caption{Qualitative examples on the Composition-1k \cite{re:DIM} test set.}
    \label{appendix:COM}
\end{figure*}

\begin{figure*}[tb]
    \centering
    \includegraphics[width=\linewidth]{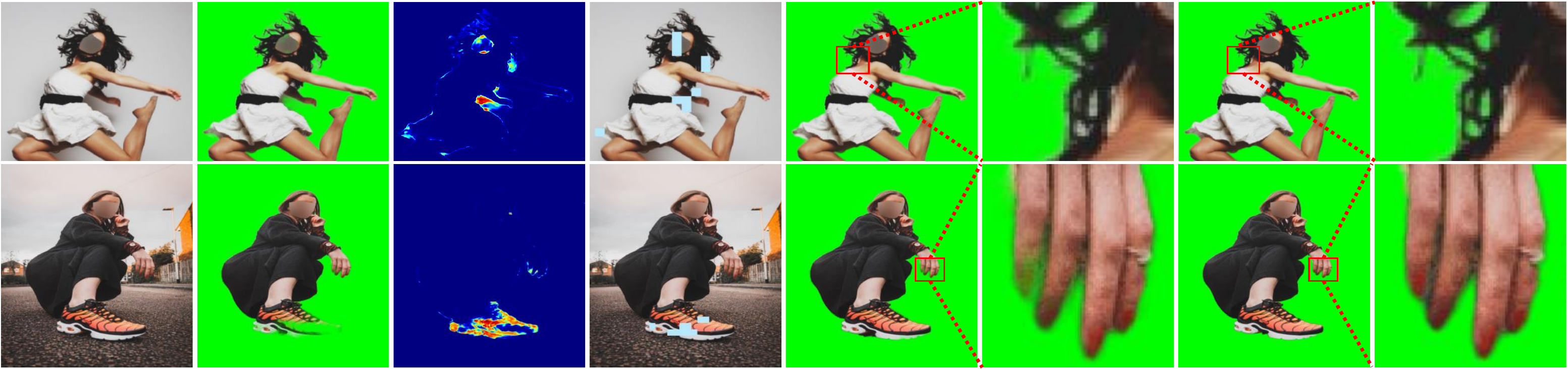}
    \caption{Visualization of step-by-step results in dugMatting. From left to right are input image, initial prediction, epistemic uncertainty, user map, prediction after interaction, prediction after refinement, respectively.}
    \label{fig:interactive}
\end{figure*}

\begin{figure}[t]
    \centering
    \begin{center}
    \subfloat[]{\includegraphics[width=0.5\linewidth]{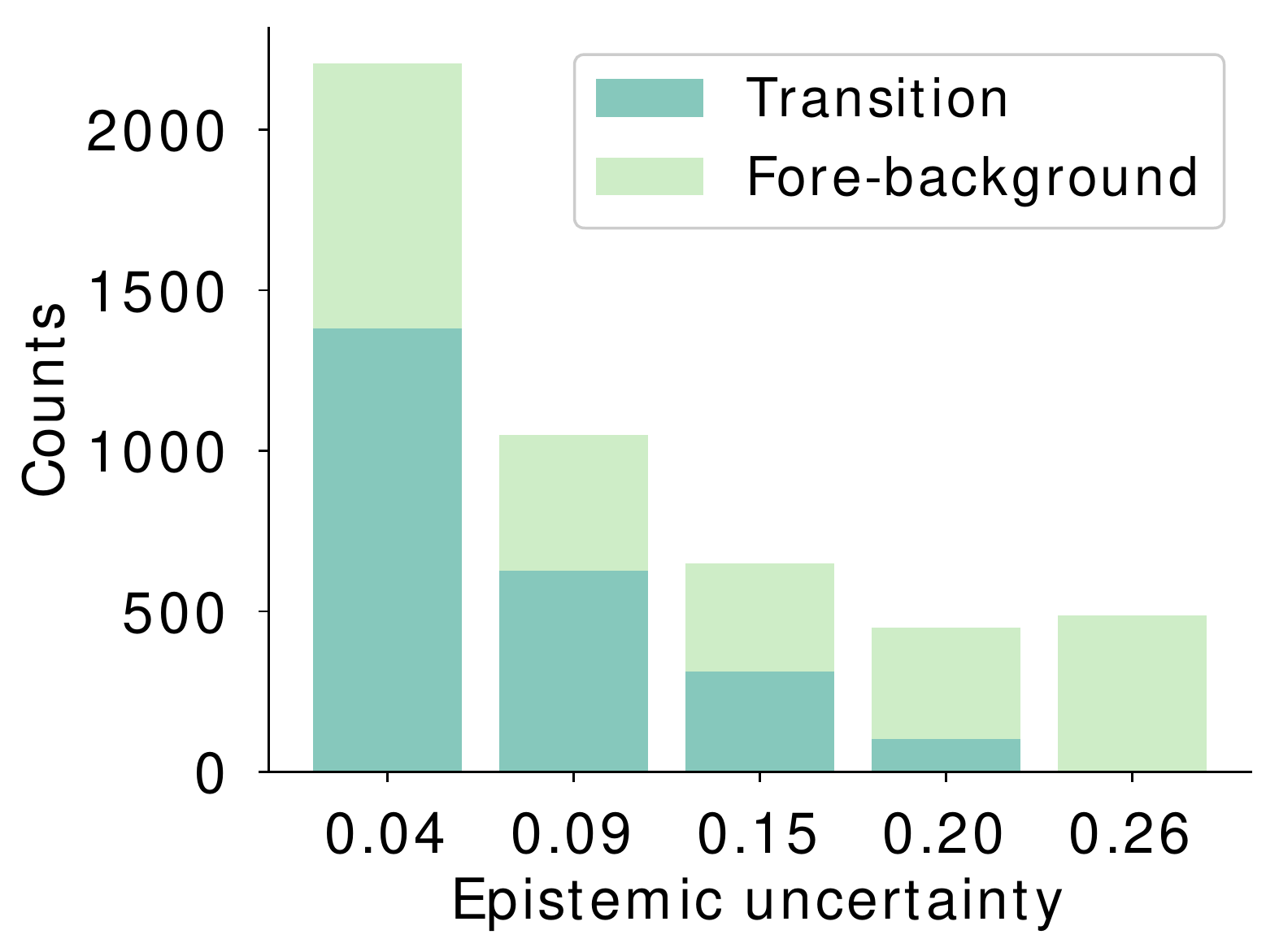}}
    \subfloat[]{\includegraphics[width=0.5\linewidth]{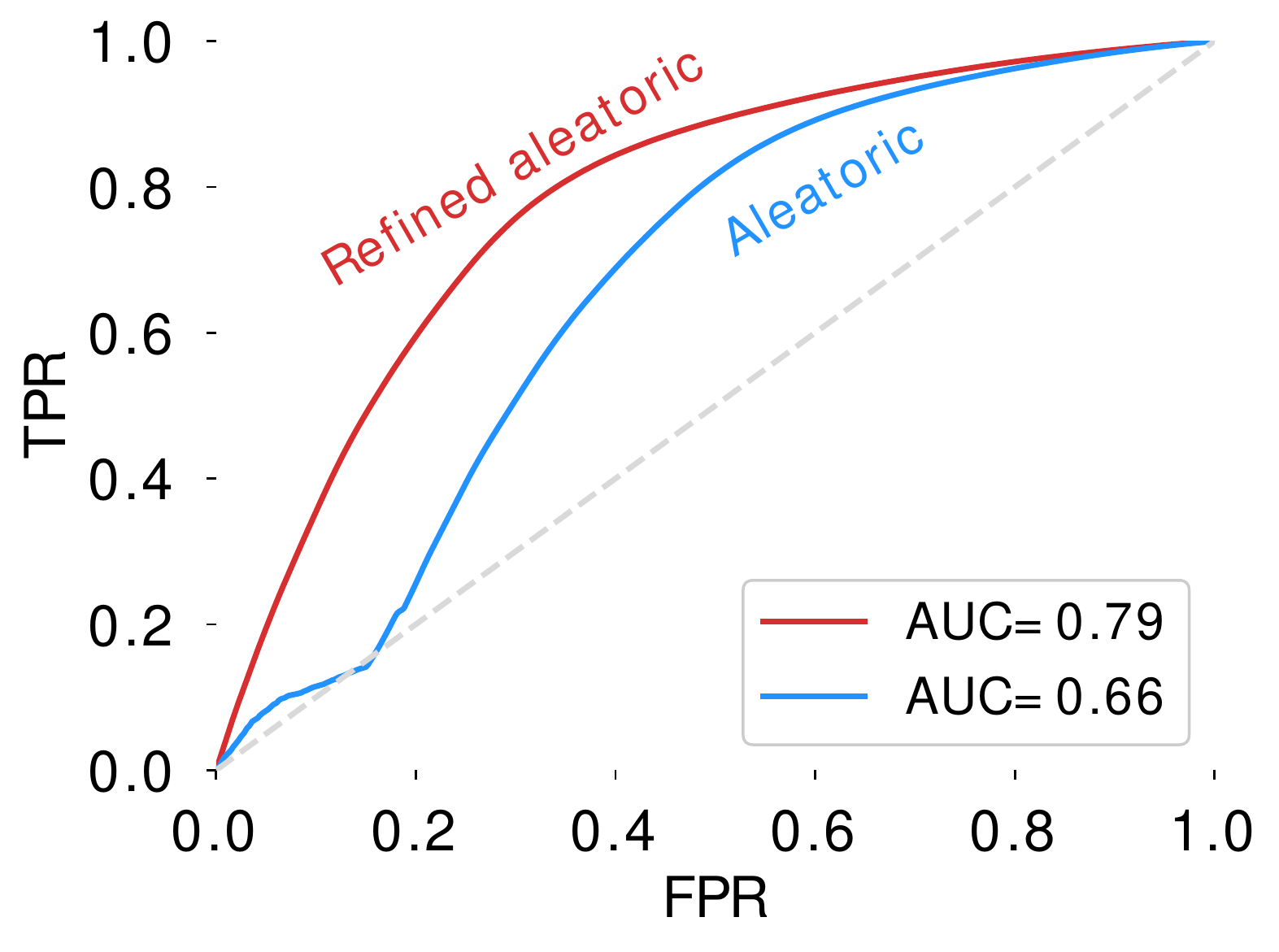}}
    \end{center}
    \caption{The correlation regions of decomposed uncertainties. The proportion of foreground and background regions is higher in high epistemic uncertainty. ROC of the obtained regions and the real transition regions, refined aleatoric uncertainty-based algorithm achieves better performance.}
    \label{fig:un_concern}
\end{figure}
\begin{figure}[t]
    \centering
    \includegraphics[width=0.9\linewidth]{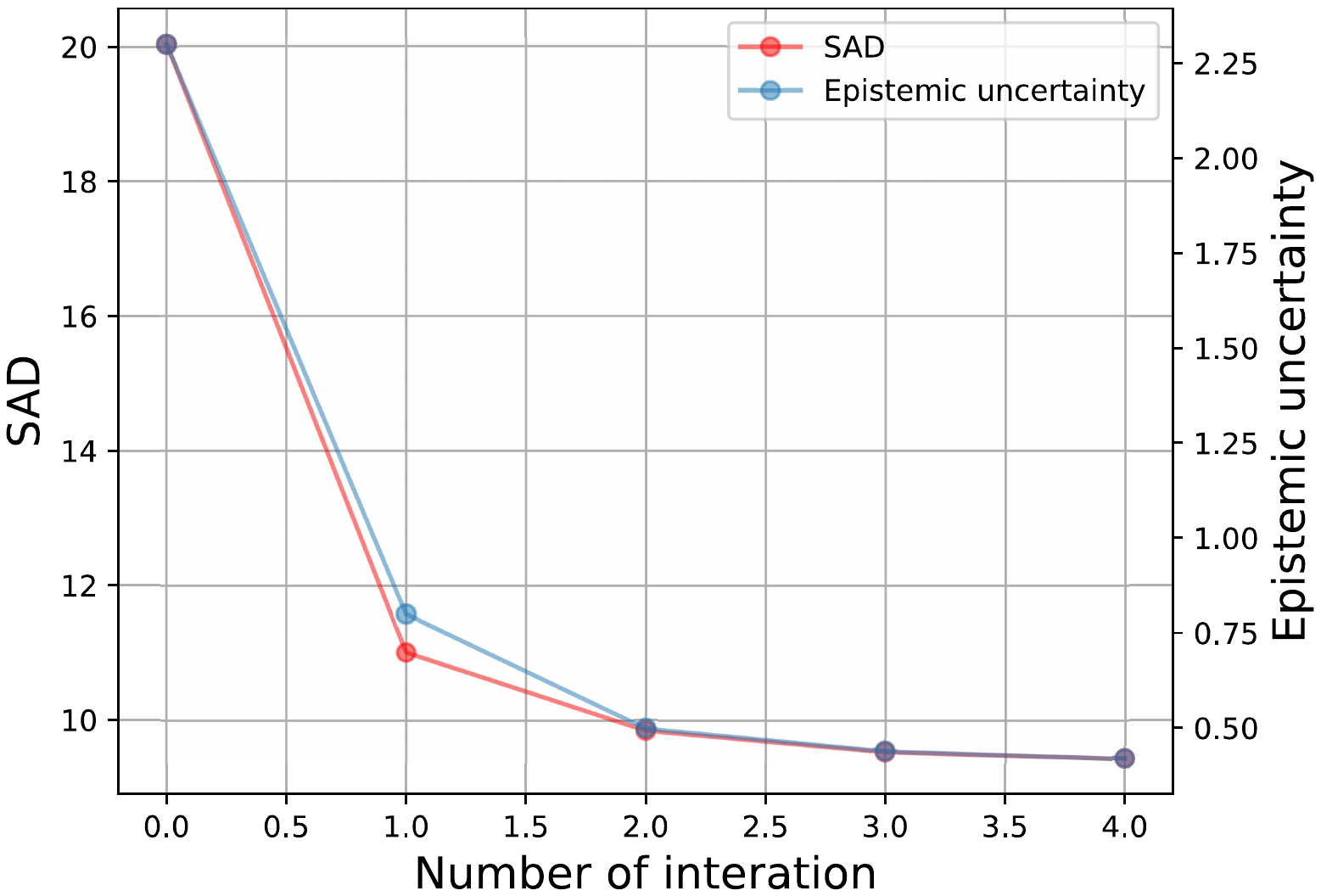}
    \caption{SAD and epistemic uncertainty at different number of interaction.}
    \label{fig:interaction_nums}
\end{figure}

\textbf{Visual Comparison with State-of-the-art Methods.}
In \cref{appendix:COM}, we visualize some results for intuitive comparison. Although dugMatting uses a weaker prior, the results is comparable to other trimap-based methods. In addition, benefiting from modeling data noise, dugMatting produces a matte that is more uniform and smooth. For instance, the ground truth of the second example has some local opacity mutations that do not occur in the real world, but dugMatting also achieves a smooth outcome.



\textbf{Visualization of Step-by-step Results in dugMatting.}
\cref{fig:interactive} visualizes the step-by-step results of our dugMatting. Our interaction can effectively improve the incorrect matting regions, and our refinement module can improve the details. The reason is that external knowledge by user interaction significantly reduces epistemic uncertainty, complementing the unlearned foreground and background patches. Meanwhile, our refinement of modeling high-frequency noise reduces the aleatoric uncertainty, enhancing the robustness in patches containing more details.



\textbf{Uncertainty Evaluation.}
We evaluate the uncertainty from two aspects. The first one is to verify the region proposal of our interaction and refinement, while the second one is to validate the ability of uncertainty estimation which is detailed in \cref{appendix:Unes}. As shown in \cref{fig:un_concern}, there are much higher proportion of foreground and background regions with large epistemic uncertainty (a). Thus selecting patches with top $K$ patch-levels epistemic uncertainty enables the user to concentrate on the annotation of foreground and background. We further evaluate the ROC curve between the regions obtained by two strategies and the real transition (b). Our refined aleatoric uncertainty-based algorithm significantly improves the AUC, demonstrating the refined aleatoric uncertainty can improve more details.

\begin{table}[tbh]
\centering
\caption{Ablation study (SAD$\downarrow$) of the NIG distribution and the proposed module on the P3M-500-P dataset.}
\label{tab:ablation}
\resizebox{\linewidth}{!}{
\begin{tabular}{c|c|c|c}
\hline
Method&  Original&  w/ NIG & w/ NIG \& Module  \\ \hline
SHM \cite{re:SHM} &  26.84&  24.65&  \textbf{21.43}\\
U$^2$Net \cite{re:u2net}&  73.48& 69.76 & \textbf{60.21} \\
MODNet \cite{re:modnet}&  23.86&  20.04 & \textbf{18.15}\\
GFM \cite{re:gfm}&  12.90&  10.89 & \textbf{9.25}\\
P3MNet \cite{re:P3M}&  12.73&  12.03 & \textbf{10.38}\\\hline
\end{tabular}
}
\end{table}

\begin{table}[tbh]
\centering
\caption{Ablation study (SAD$\downarrow$) on our refined module on the P3M-500-P dataset. Baseline uses the original trimap-free methods.}
\label{tab:ablation_module}
\resizebox{\linewidth}{!}{
\begin{tabular}{c|c|c|c}
\hline
Method&  Baseline \cite{re:modnet}&  Gaussian & Module (our)\\ \hline
SAD$_f$ &  3.69&   3.36& \textbf{3.36}\\
SAD$_b$ &  6.46&  6.55 & \textbf{6.23}\\
SAD$_t$ &  9.88&  8.75 & \textbf{8.55}\\
Aleatoric &  0.0021&  0.0015 & \textbf{0.0013}\\\hline
\end{tabular}
}
\end{table}

\subsection{Ablation Study}
In this subsection, we first investigate the proposed components and then independently analyze our plug-and-play module. Furthermore, we perform additional experiment to investigate the hyper-parameter of interaction numbers.

\textbf{The Effectiveness of Each Component.} We first evaluate the uncertainty integration in matting, i.e., replacing the deterministic output with a Normal-Inverse-Gamma distribution, and then adding the proposed plug-and-play module. As shown in~\cref{tab:ablation}, both NIG distribution and our refinement module can improve the matting performance over original methods, demonstrating the efficacy of the key components in dugMatting.


\textbf{The Effectiveness of Reducing Aleatoric Uncertainty.} Following \cite{re:reduceAl}, we compare the augmentation of our module and a Gaussian noise. The variance of Gaussian noise is fixed, determined by the average aleatoric uncertainty of all pixels. The augmentation of our module also belongs to a Gaussian noise, but the variance is dynamic and determined by the aleatoric uncertainty of the current pixel. The result of reducing the aleatoric uncertainty is shown in~\cref{tab:ablation_module}. The proposed module achieves the best performance, significantly decreasing the aleatoric uncertainty and improving the performance.

\textbf{The Hyper-parameter of Interaction Numbers.} As shown in \cref{fig:interaction_nums}, regardless of SAD or epistemic uncertainty, the most obvious improvement occurs in the first interaction, and the performance improvement is slight improved after the second interaction. Therefore, in order to balance the performance and interaction time, the interaction number is set as 1 unless otherwise specified.
\section{Conclusion}
In this paper, we propose a decomposed-uncertainty-guided matting (dugMatting) algorithm for both trimap-free and trimap-based matting. We first introduce epistemic uncertainty to actively propose interactive regions, which simplifies the search of difficult regions by user for trimap-based matting. Besides, we propose a plug-and-play module, which not only reduces the aleatoric uncertainty but also improves the matting details. This is exciting because it first explores different types of uncertainties in an explainable and elegant way in matting. 
Extensive experiments are conducted on natural matting and class-specific matting which validates that the existing matting methods equipped with dugMatting achieve superior performance than the original ones. It would be interesting to further explore the image structures (e.g., segments) for the goal of further computational efficiency and performance improvement. Another direction for further research is to apply the proposed dugMatting to other related domains such as interactive image segmentation.

\section*{Acknowledgements}

This work was supported by the National Key Research and Development Program of China (No. 2022YFC3302200), the National Natural Science Foundation of China (No. 61972187, 61976151), and the A*STAR Central Research Fund. The authors appreciate the comments from reviewers.


\bibliography{main}
\bibliographystyle{icml2023}

\newpage
\appendix
\onecolumn
\section{Proof}
\label{appendix:proof}
The marginal likelihood of Normal-Inverse-Gamma distribution by Type-II maximum likelihood technique is defined by
\begin{equation}
\begin{split}
    p(y|\tau)&=\int_{\zeta} p(y|\zeta)p(\zeta|\tau)\text{d}\zeta \\
             &=\int_{\sigma^2}^\infty\int_{\mu=-\infty}^\infty p(y|\mu,\sigma^2)p(\mu,\sigma^2|\tau)\text{d}\mu\text{d}\sigma^2 \\
             &=\int_{\sigma^2}^\infty\int_{\mu=-\infty}^\infty p(y|\mu,\sigma^2)p(\mu,\sigma^2|\gamma,\omega,\alpha,\beta)\text{d}\mu\text{d}\sigma^2 \\
             &=\int_{\sigma^2}^\infty\int_{\mu=-\infty}^\infty \left[ \sqrt{\frac{1}{2\pi\sigma^2}}\exp\left\{-\frac{(y-\mu)^2}{2\sigma^2}\right\}\right]\left[\frac{\beta}{\omega\alpha}\frac{\sqrt{\omega}}{\sqrt{2\pi\sigma^2}}\left(\frac{1}{\sigma^2}\right)^{\alpha+1}\exp\left\{-\frac{2\beta+\omega(\gamma-\mu)}{2\sigma^2}\right\}\right] \text{d}\mu\text{d}\sigma^2\\
             &=\int_{\sigma^2}^\infty \frac{\beta^\alpha\sigma^{-3-2\alpha}}{\sqrt{2\pi}\sqrt{1+1/\omega}\Gamma(\alpha)}\exp\left\{-\frac{2\beta+\frac{\omega(y-\gamma)^2}{1+\omega}}{2\sigma^2}\right\}\text{d}\sigma^2 \\
             &=\frac{\Gamma(1/2+\alpha)}{\Gamma(\alpha)}\sqrt{\frac{\omega}{\pi}}(2\beta(1+\omega))^\alpha\left(\omega(y-\gamma)^2+2\beta(1+\omega)\right)^{-(\frac{1}{2}+\alpha)}\\
             &=St\left(y;\gamma,\frac{\beta(1+\omega)}{\omega\alpha},2\alpha\right). \nonumber
\end{split}
\end{equation}
Maximizing the likelihood as \cref{eq:NLL} by using the standard parameterization for Student t distribution makes our model fit the data.

According to $\sigma^2\sim \Gamma^{-1}(\alpha,\beta)$, the $Var(\sigma^2)$ is derived from
\begin{equation}
    Var(\sigma^2)=\mathbb E((\sigma^2)^2)-\mathbb E((\sigma^2))^2, \nonumber
\end{equation}
where
\begin{equation}
\begin{split}
    \mathbb E((\sigma^2)^n)&=\frac{\beta}{\Gamma(\alpha)}\int_0^\infty \sigma^{n-2\alpha-2} \exp(-\beta/\sigma^2) \text{d} \sigma^2 \\
    &=\frac{\beta^\alpha}{\Gamma(\alpha)}\frac{\Gamma(\alpha-n)}{\beta^{\alpha-n}} \\
    &=\frac{\beta^n\Gamma(\alpha-n)}{(\alpha-1)\cdots (\alpha-n)\Gamma(\alpha-n)} \\
    &=\frac{\beta^n}{(\alpha-1)\cdots(\alpha-n)}, \nonumber
\end{split}
\end{equation}
For $\alpha>1$, we have
\begin{equation}
    \mathbb E(\sigma^2)=\frac{\beta}{\alpha-1}, \nonumber
\end{equation}
and for $\alpha>2$, we have
\begin{equation}
    \mathbb E((\sigma^2)^2)=\frac{\beta^2}{(\alpha-1)(\alpha-2)}. \nonumber
\end{equation}
Accordingly, we can obtain the variance as 
\begin{equation}
    Var(\sigma^2)=\frac{\beta^2}{(\alpha-1)^2(\alpha-2)}. \nonumber
\end{equation}


\section{More Details}
\subsection{Details of User Map}
\label{appendix:UM}
For the construction of user map $U$, we randomly sample $L$ patches with $15 \times 15$, where $L$ is drawn from a geometric distribution with $p=\frac{1}{6}$. The user map $U \in [-1,0,0.5,1]^{1\times H \times W}$ where foreground is 1, background is -1, transition is 0.5 and unknown is 0.
\subsection{Details of Refinement Module} Since the refinement module aims to recover the high-frequency details, we use the Naive Lite-HRNet-18 \cite{re:liteHRNet} and bilinear interpolation as the refinement module. The Naive Lite-HRNet-18 can efficiently preserve high-resolution features with only 0.7M parameters.

\section{More Experiments}
\subsection{Resource Comparison of Major Interaction}
\label{appendix:RC}
We also conduct a comparison experiment to explore the resource consumption of the major interaction methods. As shown in \cref{tab:interactiveM}, the trimap, scribble, and click methods do not require extra parameters while they need to take times between 17 and 260 seconds. In contrast, our method only takes 8 seconds and requires almost no extra parameters. The reason is that our interaction method actively proposes the interaction area based on the epistemic uncertainty, allowing the user to focus on the annotation. It significantly enhances the interaction efficiency.

\begin{table}[htb]
\centering
\caption{Comparison results of resource consuming on 10 samples of the Conposition-1K \cite{re:DIM} benchmark.}
\label{tab:interactiveM}
\resizebox{0.42\linewidth}{!}{
\begin{tabular}{c|c|c}
\hline
Interaction method&  Times&  Extra Parameters  \\ \hline
Trimap &  261s&  -\\
Mask&  234s&  -\\
scribble &  171s&  -\\
Click &  17s&  -\\
Selection (ours) &  \textbf{8s}&  0.7M\\\hline
\end{tabular}
}
\end{table}


\subsection{Uncertainty Estimation}
\label{appendix:Unes}
We evaluate the epistemic uncertainty and aleatoric uncertainty on unseen P3M-500-NP test dataset using MODNet. The input, absolute error, evaluation of epistemic uncertainty and aleatoric uncertainty are depicted in \cref{fig:un_eval}. For the evaluation of epistemic uncertainty, we use calibration curves to evaluate the estimation. Calibration curves are computed according to \cite{re:eva_epis}, and ideally follows $y=x$ to represent, for example, that a target falls in a 90\% confidence interval approximately 90\% of the time. It is observed that epistemic uncertainty matches error regions in most time. For the evaluation of aleatoric uncertainty, we can find that the aleatoric uncertainty is misestimated in some cases, and the variance of the aleatoric uncertainty can serve as an additional metric to identify these regions. Thus, it is appropriate for our strategy to utilize epistemic uncertainty to identify areas of user interaction and aleatoric uncertainty to guide the refinement of details.

\begin{figure*}[htb]
    \centering
    \includegraphics[width=\linewidth]{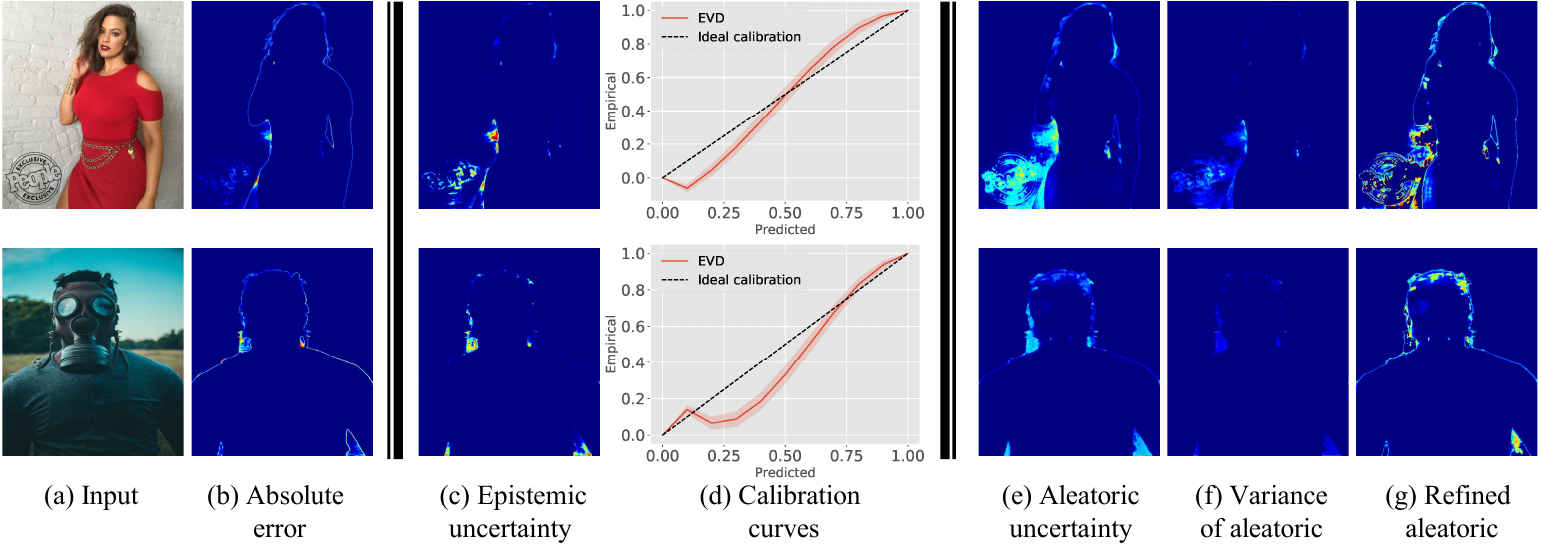}
    \caption{Uncertainty evaluation of MODNet. Epistemic uncertainty matches error regions in most time. Aleatoric uncertainty may capture erroneous transition regions, the variance of aleatoric uncertainty can help to more precisely indicate transition regions.}
    \label{fig:un_eval}
    \vspace{-5px}
\end{figure*}

\end{document}